\documentclass{ieeeaccess}
\usepackage{cite}
\usepackage{amsmath,amssymb,amsfonts}
\usepackage{algorithmic}
\usepackage{graphicx}
\usepackage{textcomp}
\usepackage{float}
\usepackage{soul}
\usepackage{subcaption}
\usepackage{multirow}
\usepackage{changepage}
\usepackage{hyperref}
\usepackage{ulem}
\makeatother
\usepackage{comment}
\usepackage{xcolor}
\usepackage{pict2e}
 
\newcommand\orcidicon[1]{\href{https://orcid.org/#1}{\usebox{\ORCIDlogo}}}
\newsavebox{\ORCIDlogo}
\savebox{\ORCIDlogo}{%
\setlength{\unitlength}{\dimexpr 1em/256\relax}%
\begin{picture}(256,256)%
  \color[HTML]{A6CE39}\put(128,128){\circle*{256}}%
  \color{white}%
  \put(78.6,199.2){\circle*{20}}%
  \moveto(70.9,176,9)\lineto(86.3,176,9)\lineto(86.3,69.8)\lineto(70.9,69.8)%
  \closepath\fillpath%
  \moveto(108.9,176.9)\lineto(150.5,176.9)%
  \curveto(190.1,176.9)(207.5,148.6)(207.5 ,123.3)%
  \curveto(207.5,95,8)(186,69.7)(150.7,69.7)%
  \lineto(108.9,69.7)%
  \closepath\fillpath%
  \color[HTML]{A6CE39}%
  \moveto(124.3,83.6)\lineto(148.8,83.6)%
  \curveto(183.7,83.6)(191.7,110.1)(191.7,123.3)%
  \curveto(191.7,144.8)(178,163)(148,163)%
  \lineto(124.3,163)%
  \closepath\fillpath%
\end{picture}%
}
\def\BibTeX{{\rm B\kern-.05em{\sc i\kern-.025em b}\kern-.08em
T\kern-.1667em\lower.7ex\hbox{E}\kern-.125emX}}
\begin{document}
\history{Date of publication xxxx 00, 0000, date of current version xxxx 00, 0000.}
\doi{10.1109/ACCESS.2017.DOI}
\title{Multimodal Attention-Enhanced Feature Fusion-based Weekly  Supervised Anomaly Violence Detection}
\author{\uppercase{Yuta Kaneko \authorrefmark{1}}, \uppercase{Abu Saleh Musa Miah \orcidicon{0000-0002-1238-0464} \authorrefmark{1},\IEEEmembership{ Member, IEEE}},
\uppercase{Najmul Hassan \orcidicon{0009-0000-6499-1825} \authorrefmark{1}, \IEEEmembership{Graduate Member, IEEE}},  \uppercase{HYOUN-SUP LEE \authorrefmark{2}},  \uppercase{SI-WOONG JANG \authorrefmark{3}}, \uppercase{Jungpil Shin \orcidicon{0000-0002-7476-2468} \authorrefmark{1},\IEEEmembership{Senior Member, IEEE}}}

\address[1]{School of Computer Science and Engineering, The University of Aizu, Aizuwakamatsu, Japan (e-mail: musa@u-aizu.ac.jp)}
\address[2]{Department of Applied Software Engineering, Dongeui University, Busanjin-Gu, Busan 47340, Republic of Korea}
\address[3]{Department of Computer Engineering, Dongeui University, Busan 47340, Republic of Korea}


\markboth
{Author \headeretal: Preparation of Papers for IEEE TRANSACTIONS and JOURNALS}
{Author \headeretal: Preparation of Papers for IEEE TRANSACTIONS and JOURNALS}
\markboth
{....}
{This paper is currently under review for possible publication in IEEE Access.}
\corresp{Corresponding author: Jungpil Shin (jpshin flu-aizu.ac.jp) and Si-Woong Jang (swjang@deu.ac.kr).}
\begin{abstract}
Weakly supervised video anomaly detection (WS-VAD) is a crucial area in computer vision for developing intelligent surveillance systems. Researchers are actively working on WS-VAD systems by assessing anomaly scores, but challenges persist due to ineffective feature extraction from unimodal approaches. Additionally, limited research on multimodal datasets has led to unsatisfactory performance accuracy. To address the challenges, We propose a multimodal attention-enhanced feature fusion-based system for weakly supervised anomaly detection. This system uses three feature streams: RGB video, optical flow, and audio signals, where each stream extracts complementary spatial and temporal features using an enhanced attention module to improve detection accuracy and robustness. In the first stream, we employed an attention-based, multi-stage feature enhancement approach to improve spatial and temporal features from the RGB video where the first stage consists of a ViT-based CLIP module, with top-k features concatenated in parallel with I3D and Temporal Contextual Aggregation (TCA) based rich spatiotemporal features. The second stage effectively captures temporal dependencies using the Uncertainty-Regulated Dual Memory Units (UR-DMU) model, which learns representations of normal and abnormal data simultaneously, and the third stage is employed to select the most relevant spatiotemporal features. The second stream extracted enhanced attention-based spatiotemporal features from the flow data modality-based feature by taking advantage of the integration of the deep learning and attention module. The audio stream captures auditory cues using an attention module integrated with the VGGish model, aiming to detect anomalies based on sound patterns. These streams enrich the model by incorporating motion and audio signals often indicative of abnormal events undetectable through visual analysis alone. The concatenation of the multimodal fusion leverages the strengths of each modality, resulting in a comprehensive feature set that significantly improves anomaly detection accuracy and robustness across three datasets. The extensive experiment and high performance with the three benchmark datasets proved the effectiveness of the proposed system over the existing state-of-the-art system. 
\end{abstract}
\titlepgskip=-15pt
\maketitle
\section{Introduction}
\label{sec1}
In video anomaly event detection (VAED), three dominant paradigms prevail: supervised, unsupervised, and weakly supervised approaches. The supervised paradigm noted for its impressive performance \cite{liu2019exploring}, relies heavily on meticulously annotated video frames categorizing normal and abnormal events \cite{sharif2023deep}.
The Variational Autoencoder Decoder (VAD), when used in an unsupervised way, often does not perform well because it struggles to fully understand anomalies and fails to recognize the different types of normal behaviours \cite{chandola2009anomaly}. To improve this, weakly supervised approaches have become popular for VAED. These methods use video-level labels instead of needing detailed frame-by-frame annotations, which makes them more cost-effective and still competitive in performance \cite{zhong2019graph,zaheer2020claws}.
Recently, Weakly Supervised VAED (WVAED) has become a major area of research \cite{zhong2019graph,zaheer2020claws,sultani2018real,zhang2019temporal,ullah2021weakly,zhu2019motion,lv2021localizing,hassan2018temporal,purwanto2021dance,miah2023dynamic,miah2024spatial_paa,miah2024_multiculture}. WVAED is often treated as a Multiple Instance Learning (MIL) problem \cite{sultani2018real}. In WVAED, models compare the spatiotemporal features of normal and abnormal events to detect anomalies. In MIL, a video is considered as a "bag" with many snippets. Negative bags contain only normal snippets, while positive bags have both normal and abnormal snippets without specific labels for when anomalies occur. MIL assumes that negative bags only have negative instances, while positive bags have at least one positive instance, though the exact labels are not provided \cite{carbonneau2018multiple}. WVAED generally performs better than unsupervised methods because it can better distinguish between normal and abnormal behaviours \cite{liu2023generalized}. However, the presence of normal snippets in positive bags can make it difficult to clearly identify anomalies at the snippet level. To address this issue, researchers are increasingly using MIL frameworks \cite{sultani2018real,zhang2019temporal,wu2020not,RTFM_tian2021weakly,joo2212clip}.



One of the most challenging tasks of the WVAED is to deal with the diverse anomalies present in a single video, including short-term motion-only, appearance-only, and audio-only anomalies [6]. 
The existing anomaly detection can categorized on various modalities: Unimodal modal methods \cite{ji20123d,carreira2017quo, shao2021temporal_TCA,yu2022tca,RTFM_tian2021weakly,pu2023learning-TCA_PEL,URDMU_zh, sharif2023cnn_CNN-ViT}, Binary modal methods including RGB+FLow, or RGB+Audio \cite{wu2020not,pang2020self,ACF_wei2022look,zhang2023exploiting,9712793,yu2022modality} and Multimodal methods including RGB+FLow+Audio \cite{9926192}. 
Many current methods use backbone architectures like C3D \cite{ji20123d} and I3D \cite{carreira2017quo}, originally pre-trained for action recognition in both unimodal and multimodal based Visual Activity Detection (VAD) due to domain differences \cite{hassan2024deep_har_miah,10624624_lstm,mallik2024virtual}. To improve feature effectiveness Contrastive Language-Image Pretraining (CLIP), have used by researchers  \cite{patashnik2021styleclip,vo2023aoe,kashu2022vltint} for using visual features from vision transformers (ViT) pre-trained with CLIP, which offer better scene representation. The success of WVAED approaches using Multiple Instance Learning (MIL) depends heavily on the quality of these pre-trained features and processing videos frame by frame or in short clips limits capturing long-range context. To improve the temporal features, Shao et al. introduced the Temporal Context Aggregation (TCA) framework  \cite{shao2021temporal_TCA,yu2022tca}, which uses self-attention mechanisms to integrate long-range temporal context \cite{Farhan_attention_miah} where they refined features using contrastive learning to reduce evaluation loss. Tean et al. improved TCA with a MIL loss approach, achieving AUC scores of 84.30\% for UCF-crime and 97.21\% for Shanghai Tech \cite{RTFM_tian2021weakly}. Pu et al. further improved TCA by using Progressive Error Learning (PEL) and integrating semantic priors, achieving AUC rates of 86.76\%, 85.59\%, and 98.14\% for UCF-crime, XD-Violence, and Shanghai Tech datasets \cite{pu2023learning-TCA_PEL}. They later introduced Uncertainty Regulated Dual Memory Units (UR-DMU) for temporal feature extraction, initially achieving 86.97\% and 94.02\% accuracy on UCF-crime and XD-violence datasets \cite{URDMU_zh}. Sharif et al. enhanced this with a dual-stream method combining CNN-based I3D and ViT-based CLIP features, reaching AUC scores of 88.97\% and 98.66\% for UCF-crime and Shanghai Tech \cite{sharif2023cnn_CNN-ViT}. More recently, Shin et al. developed a graph and general network-based anomaly detection system \cite{10510436_kaneko_anomaly} by using recent deep learning technologies. 
However, the model is unsuitable for real-time deployment due to its less effective features and the overlooked graph-based and spatial feature enhancements.  
In the UR-DMU \cite{URDMU_zh}, TCA and I3D-CLIP have explored various methods for enhancing video anomaly detection. UR-DMU \cite{URDMU_zh} focused on graph-based feature enhancement but didn't address time-varying enhancements. Conversely, TCA \cite{shao2021temporal_TCA,pu2023learning-TCA_PEL} and I3D-CLIP \cite{sharif2023cnn_CNN-ViT} tackled temporal enhancements but lacked extracting all possible feature types. Recently, Shin et al. developed a graph and general network-based anomaly detection system \cite{10510436_kaneko_anomaly}. The mentioned unimodal-based system still faces challenges in achieving good performance accuracy due to a lack of feature effectiveness because it is very difficult to capture the exact scenario of the anomaly using an unimodal data recording system due to the lack of sufficient crucial information, which could make it difficult to assess anomalies accurately. More explanation is that model performance for anomaly detection is influenced not only by the spatial contextual features of the object but also by the motion within temporal contextual features and multi-modal fusion, which is also significantly important [7]. Some researchers have been working to develop binary modal dataset fusion-based anomaly detection systems, including RGB+Audio data modalities \cite{wu2020not,pang2020self,ACF_wei2022look,zhang2023exploiting,9712793,yu2022modality}. However, their performance accuracy is not satisfactory, and few researchers have been working to develop multimodal dataset fusion-based anomaly detection systems \cite{9926192}. Moreover, some multimodal systems are constructed by multi-modal features with mutual losses, and some researchers performed the early fusion [16]. However, the main disadvantage of the system is that these modalities are not fused at the features level. Consequently, these model feature fusions do not boost the performance due to a lack of implicitly aligning the multimodal features and sometimes may fail to take advantage of the multi-modality domain.  To overcome the challenges, we proposed a weekly supervised multi-modal attention-enhanced feature fusion-based anomaly detection system.   Our approach integrates CNN and ViT-based pre-trained features, attention-based spatial-temporal features enhancement from their modalities, and spatial feature enhancements to improve anomaly detection rates effectively.

The main contributions of the proposed model are given below: 
\begin{itemize}
\item \textbf{RGB Video Stream}:
\begin{itemize}
    \item \textbf{ViT-based CLIP Module}: This branch of the RGB stream utilizes a ViT-based CLIP module to select top-k features, capturing complex visual semantics and contextual information.
    \item \textbf{CNN-based I3D Module with TCA}: The second branch leverages a CNN-based I3D module integrated with the Temporal Contextual Aggregation (TCA) mechanism to extract rich spatiotemporal features.
    \item \textbf{UR-DMU Based Feature Processing}: The combined features are processed through the Uncertainty-Regulated Dual Memory Units (UR-DMU) model, which employs GCN and GL-MHSA modules to capture video associations. Feature reduction is achieved via a multilayer perceptron (MLP), producing the final feature representation for the RGB stream.
\end{itemize}
\item \textbf{Flow Data Modality Stream}: In this stream, first we computed the motion fow from the RGB consequence frames then fed into the I3D module to capture both spatial and temporal information that also highlighting scene dynamics crucial for detecting anomalies related to unusual movements. The motion features are refined through an MLP and subsequently fed into a Transformer to capture long-range dependencies and temporal patterns, resulting in the final flow stream features.

\item \textbf{Audio Stream}: The third stream extracts features from audio signals using a Transformer applied to VGGish-extracted features. This approach captures critical audio cues for anomaly detection that may not be visible in the visual data. The VGGish model processes audio inputs into detailed feature representations, which the Transformer further enhances by capturing temporal dependencies and contextual relationships, allowing for the precise identification of subtle audio anomalies, complementing the visual streams.
\item \textbf{Gated Feature Fusion with Attention Module and Classification}: Features from all three streams are concatenated using a gated feature fusion mechanism with an attention module, producing a comprehensive final feature set for the classification module. The classifier then predicts snippet-level anomaly scores. During training, these scores are aggregated into bag-level predictions to identify high activations in anomalous cases.
\item \textbf{Comprehensive Evaluation}: Extensive experiments on the XD-Violence dataset and two other benchmark datasets demonstrate that our method outperforms state-of-the-art approaches, achieving significant improvements in anomaly detection performance.
\end{itemize}

\section{Literature Review}\label{sec2}
The methodologies employed in  WVAED rely on video-level labels, adhering consistently to the MIL framework Sultani et al.\cite{sultani2018real}. Under the MIL approach, a regression model is trained via WVAED, assuming that the highest score among positive instances exceeds that of negative instances, thereby assigning scores to video snippets. Previous studies  \cite{sultani2018real}, \cite{RTFM_tian2021weakly}, \cite{zhang2019temporal},  \cite{zhong2019graph}, \cite{zhu2019motion} have integrated pre-trained CNN models into their experimental setups.
Sultani et al.\cite{sultani2018real} curated pre-annotated normal and abnormal video events at the video level, establishing the widely used UCF-Crime dataset for anomaly detection within a weakly supervised framework, involved extracting C3D features \cite{tran2015learning} from video segments and subsequently employing a ranking loss function to train a fully connected neural network (FCNN) \cite{miah2022bensignnet,computers12010013_multistage_musa,electronics12132841_multistream,shin2024korean_ksl0}.
The objective of this function was to calculate the loss between the highest-scoring ranked examples within the positive and negative bags. Tian et al. \cite{RTFM_tian2021weakly}  introduced a model for WVAED, leveraging feature extractors such as C3D \cite{tran2015learning}, and I3D \cite{carreira2017quo}. They argued that by selecting the top three features based on their magnitude, a more distinct differentiation could be achieved between normal and abnormal videos (AVs). Particularly, in cases where multiple abnormal snippets exist within an anomalous video, the average feature magnitude of the anomalous video exceeds that of normal videos (NVs).
Zang et al. \cite{zhang2019temporal} proposed a model using temporal convolution networks (TCN) to extract C3D features from positive and negative video segments. They trained the network to discern between adjacent segments, employing inner and outer bag ranking losses to train their model with two branches of an FCNN \cite{shin2023korean_ksl1,shin2024japanese_jsl1,10360810_miah_ksl2,miah2023dynamic_mcsoc,miah2023skeleton_euvip}. This approach focused on the highest and lowest-scoring segments within positive and negative bags, respectively.
Similarly, Zhong et al. \cite{zhong2019graph} and Zhu et al. \cite{zhu2019motion} developed models that concurrently trained feature-based encoders and classifiers. Zhong et al. \cite{zhong2019graph} approached WVAED as a supervised learning problem, leveraging noisy labels. Their study extensively evaluated the general applicability of their model, integrating both temporal segment networks \cite{wang2018temporal} and C3D \cite{wang2018temporal}.
Zhu et al. \cite{zhu2019motion}  introduced an attention mechanism into their MIL ranking model to capture temporal context. They demonstrated that motion information extracted by C3D \cite{tran2015learning} and I3D \cite{carreira2017quo} outperformed features derived from individual images perform pre-trained models like VGGish16 \cite{simonyan2014very} and Inception \cite{szegedy2015going,rahim2024advanced_miah}.
ViT-based pre-trained models can be categorized into single-stream and dual-stream architectures. In the single-stream approach, both textual and visual (or video) information are encoded within a unified transformer framework, whereas the dual-stream model employs separate encoders to handle text and image (or video) inputs independently. Notable ViT feature extractors include CLIP \cite{radford2021learning}, ViLBERT \cite{lu2019vilbert}, VisualBERT \cite{li2019visualbert}, and data-efficient CLIP \cite{li2021supervision}.
Addressing the WVAED problem, Joo et al. \cite{joo2212clip} recently introduced a temporal self-attention framework assisted by CLIP, conducting experiments on publicly available datasets to validate their end-to-end WVAED model. Li et al. \cite{li2022self} proposed a MIL network based on transformers to compute anomaly scores for both entire videos and video snippets, utilizing video-level anomaly probabilities during inference to stabilize snippet-level anomaly scores. Lv et al. \cite{lv2023unbiased} developed an unbiased MIL approach that trains a fair anomaly classifier alongside a tailored representation specifically designed for WVAED.
In current practice, CNN and ViT models are typically applied independently \cite{shin2023korean_ksl1,miah2024_multiculture,rahim2020hand}. To integrate the strengths of both CNN- and ViT-based pre-trained models, researchers have devised architectures like CNN-ViT-TSAN supported by sMIL. This framework aims to offer a diverse array of models tailored to addressing challenges in  WVAED.
The primary limitation and challenges of previous studies lie in their approach to processing videos frame by frame or in short clips, which limits their ability to capture long-range semantic contextual information effectively. To address this challenge, \cite{shao2021temporal_TCA,yu2022tca} proposed a TCA framework for video representation learning. This innovative method integrates long-range temporal context into frame-level features using self-attention mechanisms \cite{shao2021temporal_TCA,yu2022tca}. They employed contrastive learning to mitigate loss or error rates during evaluation.
To further enhance TCA features, they utilized Robust TCA features alongside a MIL loss calculation approach \cite{RTFM_tian2021weakly}. Their approach achieved notable results with reported AUC values of 84.30\% for the UCF-Crime dataset and 97.21\% for the Shanghai Tech dataset.
Pu et al.\cite{pu2023learning-TCA_PEL} sought to improve AUC rates by focusing on feature effectiveness. They employed TCA to enhance long-range dependencies and introduced  PEL instead of contrastive learning to enhance correct prediction rates by reducing errors \cite{pu2023learning-TCA_PEL}. Their method incorporated a Multi-Layer Perceptron (MLP) with PEL for feature reduction and  CC for classification. PEL integrates semantic priors through knowledge-based prompts to enhance recognition rates and discriminative capacity, ensuring high separability between anomaly subclasses. They reported impressive AUC rates of 86.76\%, 85.59\%, and 98.14\% for the UCF-Crime, XD-Violence, and Shanghai Tech datasets, respectively. This underscores the effectiveness of their approach in improving anomaly detection performance.
To improve performance accuracy rate, Zhao et al. introduced UR-DMU, focusing on temporal feature extraction using graph-based transformers via the I3D backbone \cite{URDMU_zh}. They achieved 86.97\% and 94.02\% accuracy on UCF-crime and XD-violence datasets, respectively. Sharif et al. later proposed a two-stream approach for temporal feature enhancement, combining CNN-based I3D and ViT-based Clip features \cite{sharif2023cnn_CNN-ViT}. They reported 88.97\% and 98.66\% AUC for UCF crime and Shanghai tech datasets but faced challenges in real-time deployment due to feature effectiveness issues. Both studies do not integrate graph-based and spatial feature enhancements. UR-DMU \cite{URDMU_zh} incorporated graph-based features but lacked time-varying enhancements, while TCA \cite{shao2021temporal_TCA,pu2023learning-TCA_PEL} and I3D-CLIP \cite{sharif2023cnn_CNN-ViT} addressed temporal aspects but overlooked complementary features. Inspired by these gaps, we propose a novel anomaly detection system leveraging multi-stage graphs and deep learning feature enhancements. Our approach integrates CNN and ViT pre-trained features, temporal enhancements, graph-based temporal features, and spatial feature enhancements to optimize anomaly detection performance.

\section{Background and Theoretical Foundation}
This section outlines the theoretical foundation and background of our deep learning framework for video-based anomaly detection. We detail the mathematical representations underpinning the model, which integrates a range of advanced techniques. These include a 3DCNN, ViT-based CLIP module for visual feature extraction, CNN-based I3D enhanced with Temporal Context Aggregation (TCA), and Uncertainty-Resilient Dual Memory Units (UR-DMU) with Global/Local Multi-Head Self Attention (GL-MHSA) and Transformer. Paired with a multilayer perceptron (MLP), these elements create a powerful system for accurately detecting anomalies in video data.

\subsection{CNN Model Construction}  \label{theory:CNN}
Convolutional Neural Networks (CNNs) are effective at extracting spatial features from images. For an input image \( I \), a CNN generates a feature map \( \mathbf{F}^{CNN} \) as follows:

\begin{equation} \label{eq:cnn_features}
\mathbf{F}^{CNN} = \text{CNN}(I)
\end{equation}

This feature map is derived through convolution operations:

\begin{equation} \label{eq:conv_operation}
\mathbf{F}_{i,j,k} = \text{ReLU}\left( \sum_{m,n} I_{i+m, j+n} \cdot W_{m,n,k} + b_k \right)
\end{equation}

 Here, \( W \) are the convolutional filters, \( b_k \) is the bias, and ReLU introduces non-linearity. Pooling operations then reduce spatial dimensions:

\begin{equation} \label{eq:max_pooling}
\mathbf{P}_{i,j} = \max_{m,n} \mathbf{F}_{i+m, j+n}
\end{equation}

For video-based anomaly detection, 3D Convolutional Neural Networks (3D CNNs) extend these operations to three dimensions, allowing the model to capture both spatial and temporal features. The 3D convolution operation is:

\begin{equation} \label{eq:3dconv_operation}
\mathbf{F}_{i,j,k,l} = \text{ReLU}\left( \sum_{m,n,p} I_{i+m, j+n, k+p} \cdot W_{m,n,p,l} + b_l \right)
\end{equation}

Here, \( I \) is the 3D input (e.g., video frames), and \( \mathbf{F}_{i,j,k,l} \) is the resulting 3D feature map. 3D pooling is similarly extended:

\begin{equation} \label{eq:3dmax_pooling}
\mathbf{P}_{i,j,k} = \max_{m,n,p} \mathbf{F}_{i+m, j+n, k+p}
\end{equation}
By capturing spatial and temporal features, 3D CNNs are ideal for video-based anomaly detection, providing an edge over 2D CNNs in dynamic environments \cite{tiwari2023comprehensive_3DCNN}.

\subsection{I3D Model Construction}  \label{theory:I3D}
To derive the Inflated 3D ConvNet (I3D) model from the 3D CNN framework, we extend the principles of 3D convolutions to handle video sequences. Given a video sequence \( V_v \) consisting of \( T_v \) snippets, the I3D model extracts features as follows:
\begin{equation} \label{eq:i3d_features}
\mathbf{F}^{I3D}_{v} = \text{I3D}(V_{v})
\end{equation}
Here, \( \mathbf{F}^{I3D}_{v} \) represents the feature matrix obtained from the I3D model, where \( T_v \) is the number of snippets and \( \aleph \) is the dimensionality of the feature vector for each snippet. The feature extraction process involves applying 3D convolutions to capture both spatial and temporal information from the video sequence:
\begin{equation} \label{eq:3d_convolution}
\mathbf{F}^{I3D}_{v}(t, f, c) = \text{Conv}_{3D}(\mathbf{V}_{v}(t, f, c))
\end{equation}
In this equation, \( \mathbf{V}_{v}(t, f, c) \) is the video input at time \( t \) and spatial location \( (f, c) \), and \( \text{Conv}_{3D} \) denotes the 3D convolution operation which explained in Section \ref{theory:CNN}.  This operation is crucial for capturing both motion and appearance cues over time in the I3D model.
The I3D model extracts features from the video sequence, effectively representing both motion and appearance cues over time. The features are computed from the \( T_v \) snippets, resulting in a feature vector \( \acute{\phi_{v_{\text{cnn}}}} \) with reduced dimensions:
\begin{equation} \label{eq:reduced_feature}
\acute{\phi_{v_{\text{cnn}}}} =  {\phi_{i}}^{T_{v}}_{i=1} \in \mathbb{R}^{T_{v} \times \acute{\aleph}} 
\end{equation}
This is derived from the original feature dimension \( \phi_{v_{\text{cnn}}} \in \mathbb{R}^{T_{v} \times \aleph} \), where \( \aleph \) is the feature dimension extracted from \( T_v \) snippets.

By inflating 2D convolutions into 3D, the I3D model enhances the capability of traditional CNNs to analyze video data, capturing both spatial and temporal dynamics. This makes I3D particularly suitable for video-based anomaly detection, where understanding motion and appearance over time is crucial.

\subsection{Multi-Layer Perceptron (MLP)}  \label{theory:MLP}
A Multi-Layer Perceptron (MLP) is a neural network composed of fully connected layers, where each neuron in one layer is connected to every neuron in the next layer. For an input feature vector \( \mathbf{x} \), the MLP computes the output as:
\begin{equation} \label{eq:mlp}
\mathbf{y} = \sigma(\mathbf{W}_2 \cdot \sigma(\mathbf{W}_1 \cdot \mathbf{x} + \mathbf{b}_1) + \mathbf{b}_2)
\end{equation}
Here, \( \mathbf{W}_1 \) and \( \mathbf{W}_2 \) are weight matrices, and \( \mathbf{b}_1 \) and \( \mathbf{b}_2 \) are bias vectors. 
While 3D CNNs capture spatial and temporal patterns through convolutional operations across video data, MLPs focus on dense feature mapping through fully connected layers, making them inherently different in how they process data.

\subsection{Multi-Head Self-Attention} \label{theory:mhsa}
The Multi-Head Self-Attention (MHSA) mechanism can be seen as an extension of the MLP. While an MLP uses fully connected layers to map input features to higher-dimensional spaces, MHSA extends this by allowing the model to focus on different parts of the input simultaneously. 
Given an input sequence of features \( \mathbf{F} = [F_1, F_2, \dots, F_n] \), the self-attention mechanism computes a weighted sum of the values, considering the relevance of each feature to the others:
\begin{equation}
\text{Attention}(Q, K, V) = \text{softmax}\left(\frac{QK^T}{\sqrt{d_k}}\right)V
\end{equation}
Here, \( Q, K, V \) are the query, key, and value matrices derived from the input feature vector \( \mathbf{F} \). \( d_{k} \) is the dimensionality of the key vectors.
The MHSA mechanism extends this by computing multiple attention heads in parallel:
\begin{equation}
\text{MultiHead}(Q, K, V) = \text{Concat}(\text{head}_1, \dots, \text{head}_h)W^O
\end{equation}
Where each attention head \( \text{head}_i \) performs a separate attention operation:
\[
\text{head}_i = \text{Attention}(QW_i^Q, KW_i^K, VW_i^V)
\]
After the self-attention layer, the output passes through a feed-forward network (FFN), similar to an MLP:
\begin{equation}
\text{FFN}(x) = \text{ReLU}(xW_1 + b_1)W_2 + b_2
\end{equation}
The MHSA module enhances the input feature vector \( \mathbf{F} \) by allowing the model to attend to multiple aspects of the input simultaneously, producing an enriched feature representation \( \mathbf{F}_{enhanced} \). This approach encapsulates both local and global information, making it more powerful than a traditional MLP for capturing complex dependencies in the input data.

\subsection{Temporal Context Aggregation Module (TCA)} \label{theory:TCA}
The Temporal Context Aggregation (TCA) module enhances I3D features by integrating both local and global temporal dependencies, improving anomaly detection. TCA uses self-attention mechanisms, similar to MHSA, to capture relationships across different time steps in a sequence, while also benefiting from UR-DMU's memory-based approach to distinguish between normal and anomalous data.

TCA utilizes a self-attention mechanism, akin to MHSA, to capture relationships between different time steps in a sequence. This allows the model to focus on various parts of the input sequence, similar to how MHSA operates within Transformer architectures. The self-attention mechanism in TCA is particularly crucial for understanding both short-term and long-term temporal relationships, which are essential for effective anomaly detection. We described the TCA driven formul below:
\begin{enumerate}
    \item Calculating a similarity matrix \( M \) from the I3D output \( X \) projected into a latent space:
    \begin{equation} \label{eq_sim_matrix}
        M = f_q(X) \cdot f_k(X)^\top
    \end{equation}
    \item Computing global attention \( A^g \) using softmax:
    \begin{equation} \label{eq_g_att}
        A^{g} = \text{softmax}\left( \frac{M}{\sqrt{D_h}} \right)
    \end{equation}
    \item Extracting global context features \( X^g \) by applying \( A^g \) to \( f_v(X) \):
    \begin{equation} \label{eq_g_feature_matrix}
        X^{g} = A^{g} \cdot f_v(X)
    \end{equation}
    \item Enhancing the similarity matrix with Dynamic Position Encoding (DPE):
    \begin{equation} \label{eq:dpe}
        \mathbf{G} = \exp(-|\gamma(i-j)^2 + \beta|)
    \end{equation}
    \item Calculating local attention \( A^l \) and local context features \( X^l \):
    \begin{equation} \label{eq_l_att}
        A^{l} = \text{softmax}\left( \frac{\tilde{M}}{\sqrt{D_h}} \right)
    \end{equation}
    \begin{equation} \label{eq_l_feature_matrix}
        X^{l} =  A^{l} \cdot f_v(X)
    \end{equation}
    \item Aggregating final features \( X^o \) by combining global and local features:
    \begin{equation} \label{eq:final_feature}
        X^{o} = \alpha \cdot X^{g} + (1 - \alpha) \cdot X^{l}
    \end{equation}
    \item Producing the final output \( X^c \) after normalization and concatenation:
    \begin{equation} \label{eq:tca_feature}
        X^{c} = \text{LN}(X + f_h(\text{Norm}( X^{o})))
    \end{equation}
\end{enumerate}

The TCA module effectively enhances I3D features, making it more powerful in capturing temporal dependencies for robust anomaly detection.
While the MHSA mechanism in TCA is primarily concerned with capturing temporal dependencies, the UR-DMU model goes a step further by incorporating memory units that store and distinguish between normal and abnormal data over time. The TCA's emphasis on integrating both global and local temporal information through self-attention can be seen as complementary to UR-DMU’s focus on retaining important feature representations for anomaly detection. Both modules enhance the model’s ability to differentiate between normal and abnormal behaviours, albeit through different mechanisms.

\subsection{UR-DMU} \label{theory:ur-dmu}
The Uncertainty-Resilient Dual Memory Unit(UR-DMU) model extends the capabilities of the Multi-Head Self-Attention (MHSA) mechanism by introducing dual memory units that enhance the model's ability to distinguish between normal and anomalous data.  Building on MHSA, UR-DMU incorporates dual memory units to learn regular data representations and discriminative features of anomalies simultaneously. This approach enhances the temporal dependency-capturing abilities of MHSA by adding a memory mechanism that retains critical features over time.

\paragraph{Components of UR-DMU}
UR-DMU integrates Global and Local Multi-Head Self-Attention (GL-MHSA) with a memory augmentation mechanism:
\begin{equation}
\label{eq:memory_aug}
\mathbf{S} = \sigma\left(\frac{\mathbf{X}\mathbf{M}^t}{\sqrt{D}}\right), \quad \mathbf{M}_{aug} = \mathbf{S}\mathbf{M}
\end{equation}
Here, \( \mathbf{X} \) is the feature from GL-MHSA, \( \mathbf{M} \) is the memory bank, and \( D \) is the output dimension. This mechanism allows learning from both current and historical contexts.
\paragraph{Dual Memory Loss}
UR-DMU uses a dual memory loss function, comprising multiple binary cross-entropy (BCE) losses:
\begin{equation}
\label{eq:dual_memory_loss}
\begin{split}
L_{dm} &= \text{BCE}(\mathbf{S}^n_{k;n}, \mathbf{y}^n_{n}) + \text{BCE}(\mathbf{S}^n_{k;a}, \mathbf{y}^n_{a}) \\ 
&\quad + \text{BCE}(\mathbf{S}^a_{k;n;k}, \mathbf{y}^a_{n}) + \text{BCE}(\mathbf{S}^a_{k;a;k}, \mathbf{y}^a_{a})
\end{split}
\end{equation}
Here, \( \mathbf{S}^n_{k;n} \) represents the normal memory score, \( \mathbf{y}^n_{n} = \mathbf{1} \in \mathbb{R}^N \), \( \mathbf{S}^n_{k;a} \) is the anomaly memory score, and \( \mathbf{y}^n_{a} = \mathbf{0} \in \mathbb{R}^N \). The top-K results \( S^a_{k;n;k}, S^a_{k;a;k} \in \mathbb{R}^N \) are the means along the first dimension of \( \mathbf{S}^a_{k;n} \) and \( \mathbf{S}^a_{k;a} \), respectively, where \( y^a_{n}, y^a_{a} \) are the labels with values set to 1. This function improves the model's ability to differentiate between normal and anomalous data by comparing learned features with stored memory templates.
\paragraph{Normal Data Uncertainty Learning (NUL)}
UR-DMU also integrates Normal Data Uncertainty Learning (NUL), which applies a Gaussian distribution to constrain customary data representations, adding robustness to the model in anomaly detection tasks.

\section{Proposed Method}
In our proposed model for video-based anomaly detection, we leverage a sophisticated combination of state-of-the-art technologies to enhance the accuracy and robustness of anomaly identification. Figure \ref{fig:proposed_model} demonstrates the proposed model where we used a multi-stage deep learning (DL) approach \cite{computers12010013_multistage_musa}. This study is mainly designed to extract characteristics that are more indicative of anomalies. Similar to previous work \cite{pu2023learning-TCA_PEL, URDMU_zh, joo2023cliptsa}, we extract features from each video with 5-crop augmentation for the XD-Violence dataset using pre-trained models. There are three modalities used in the dataset to extract the features from the three different streams, including  RGB video, flow, and audio signal modalities feature streams.
We divided the untrimmed video into non-overlapping snippets using a 16-frame sliding window for the RGB video data modality stream. Then, we introduce a multi-backbone framework, combining a CLIP model trained on Kinetics with an I3D model also pre-trained on Kinetics. This dual-backbone approach leverages the strengths of both architectures to enhance the feature extraction process for video anomaly detection. Subsequently, this enhanced feature set is streamlined via TCA, CNN, UR-DMU, and a two-layer Multilayer Perceptron (MLP), optimizing it for further analysis or applications which are considered as first stream features. This section of the RGB video data stream is nearly identical to that of the previous model. It loads the weights from the previous model before beginning training \cite{10510436_kaneko_anomaly}.
In the Flow data modality Stream, we first computed the fow or motion from the RGB consequence frames then fed into the I3D module to capture both spatial and temporal information. This also emphasize the dynamics within the scene, which is crucial for identifying anomalies characterized by unusual movements. These I3D based motion features are fed into an MLP module to learn a compact and informative representation. Subsequently, a Transformer processes these features to capture long-range dependencies and temporal patterns in the motion data and produce the 2nd stream features. 
The Audio Stream leverages the power of a Transformer integrated with a VGGish model to extract rich and nuanced features from audio signals. This stream is crucial for capturing anomalies detectable through unusual audio patterns that may not be evident in the visual data. The VGGish model, known for its strong performance in audio classification tasks, processes the audio input to produce a detailed feature representation. The Transformer then enhances this representation by capturing temporal dependencies and contextual relationships within the audio data. This combination allows the model to identify subtle audio anomalies with high precision, providing a robust layer of detection that complements the visual streams. Finally, we concatenated the features using the gated feature function with attention technique, which fed into the classification module in two ways: concatenated features of the RGB video and Flow modality and concatenated features of the RGB video, Flow, and Audio signal-based features.

\Figure[htp](topskip=0pt, botskip=1pt, midskip=0pt)[scale=.33]{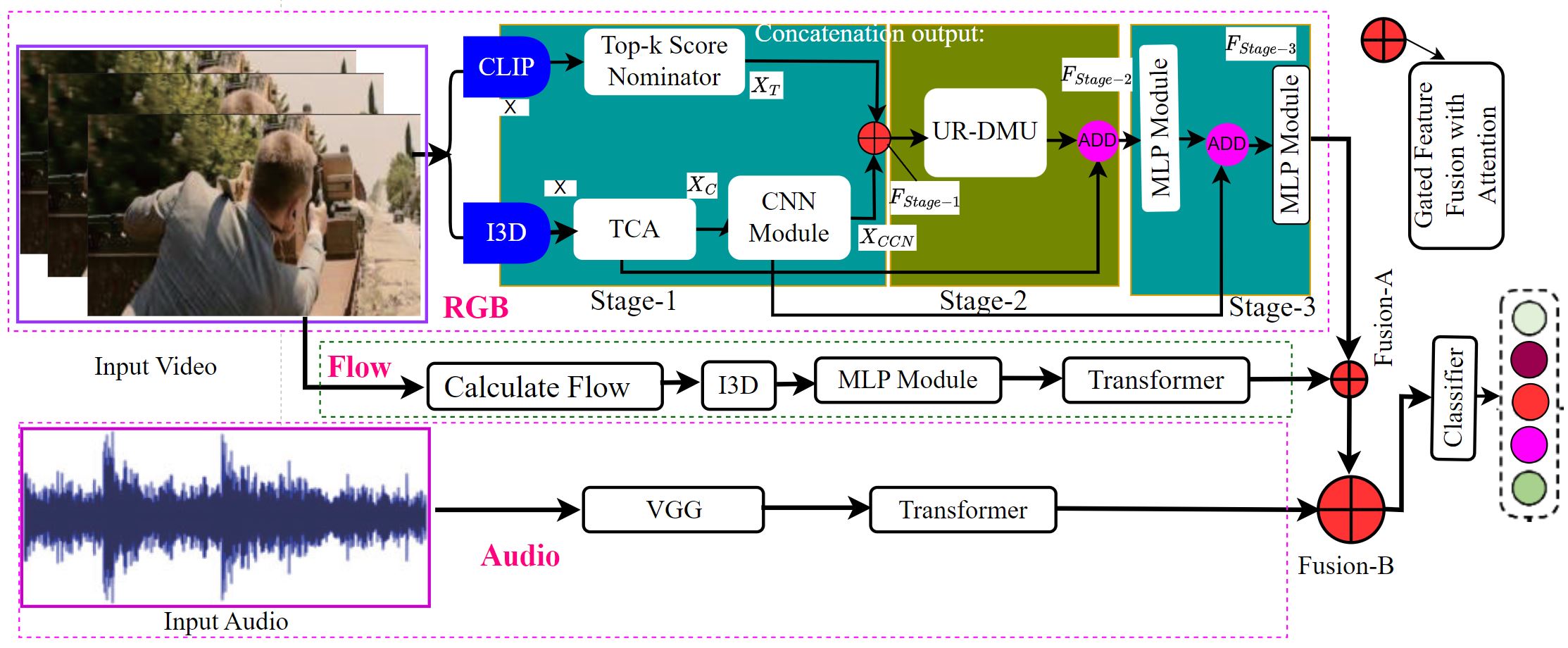}{Proposed Model\label{fig:proposed_model}}

\subsection{Peprocessing}
In WVAED, the training set consists solely of video-level labels. The set of training videos can be expressed as \( W = \{ (V_v, y_v) \}_{v=1}^w \), where each video \( V_v = \{ \text{Frame}_i \}_{i=1}^{N_v} \in \mathbb{R}^{N_v \times W \times H} \) represents a sequence of frames \( N_v \), and each frame has a width \( W \) and a height \( H \). The label of each video \( V_v \), denoted as \( y_v \in \{0, 1\} \), indicates the presence of an anomaly.
For each video, we divided it into a set of snippets, expressed as \( \{ \gamma_i \}_{i=1}^{\lfloor \frac{N_v}{\Delta} \rfloor} \), where each snippet contains an equal number of frames \( \Delta \).
In the preprocessing step, we followed the existing methodology. First, we divided the untrimmed video into non-overlapping snippets using a 16-frame sliding window \cite{pu2023learning-TCA_PEL, URDMU_zh, joo2023cliptsa}. Then, we extracted features from each sample using 5-crop augmentation for the XD-Violence dataset, utilizing pre-trained models in the initial stage \cite{pu2023learning-TCA_PEL, URDMU_zh, joo2023cliptsa}.

\subsection{RGB Data Modality Stream}
The dynamic or RGB Video we considered as the data modality of the first stream, which was constructed with three stages, which we defined as the Stage-1 initial feature, then stage-2 feature enhancement and stage-3 as feature reduction, which is described below. 

\subsubsection{Stage 1: Pretrained Model-Based Feature Extraction}
In the first stage, we introduce a multi-backbone framework, combining a CLIP model trained on Kinetics with an I3D model pre-trained on Kinetics. It is important to note that our I3D model extracted RGB and flow features. In this architecture, the I3D RGB model extracts features in 1024-dimensional space with 1024-dimensional Flow features, while the CLIP model provides feature vectors in 512 dimensions. This dual-backbone approach leverages the strengths of both architectures to enhance the feature extraction process for video anomaly detection.

\paragraph{Top-k Score Nominator Selection from CLIP Transformer Features}
The first stream is composed by integrating the CLIP pre-trained approach feature with the Top-k score nominator. Here, CLIP leverages ViTs to capture the correlation among the frames, mainly extracting the intricate internal relationship among the frames. The main concept of the CLIP model is that it is composed of a multi-backbone framework \cite{joo2212clip}. In the study, we considered \( d_{j} = \lceil \frac{\Delta}{2} \rceil \) as the middle frame of the video snippet \(\gamma_{j}\), which means we did not consider all frames simultaneously. In our study, we applied the CLIP model to \( d_{j} \) of the snippet \(\gamma_{j}\) to represent its features as \( \phi_{v_{\text{j}}} \in \mathbb{R}^{\aleph} \), where \(\aleph\) represents the feature dimension. The final feature vector is constructed as \( \phi_{v_{\text{vit}}} = \{ \phi_{j} \}_{j=1}^{T_{v}} \in \mathbb{R}^{T \times \aleph} \) \cite{sharif2023cnn_CNN-ViT}. It produces a separate feature for each video, and each feature's dimension size is 512.
The output of the CLIP model is fed into the K-Score Selection Module, shown in Figure \ref{fig:topk_score}. This module selects the most relevant video snippets based on the top-k score nominator, as described by Joo et al. \cite{joo2023cliptsa}. The process involves cloning the CLIP model output, adding Gaussian noise, and calculating the magnitude. The top K scores are then selected to focus on the most significant parts of the video.

\Figure[htp](topskip=0pt, botskip=1pt, midskip=0pt)[scale=.40]{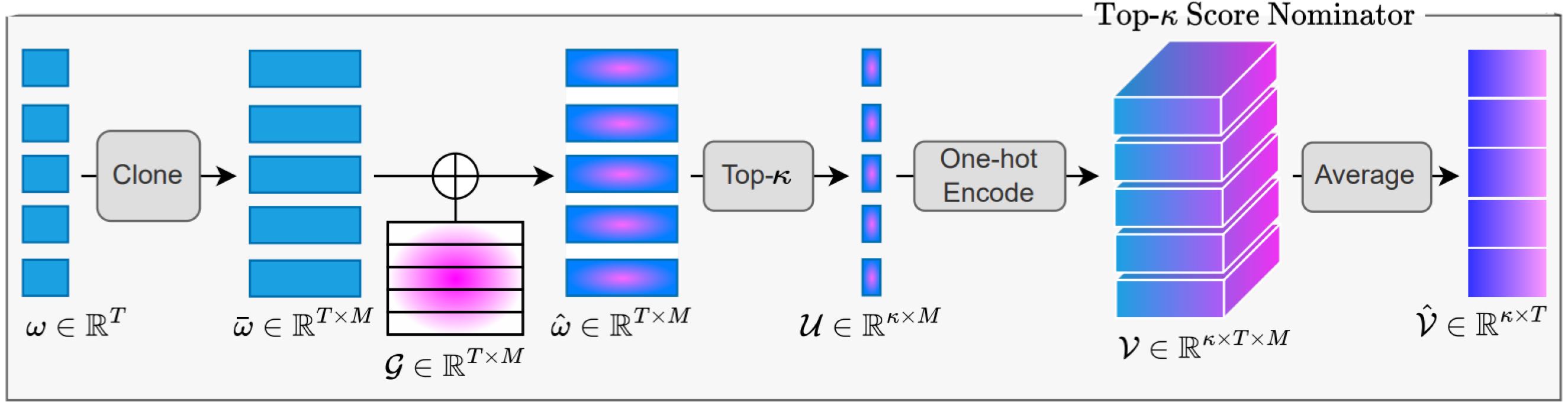}{Internal structure of the Top K-Score Selection module \cite{joo2023cliptsa,10510436_kaneko_anomaly}\label{fig:topk_score}}
\paragraph{Spatial-Temporal Feature Enhancement By With I3D and TCA with CNN:}
In the second stream, we first employed I3D \cite{carreira2017quo} to enhance the frame-wise spatial feature for the anomaly video data, then applied TCN to improve the temporal feature and generate spatial-temporal features. After that, it fed into the CNN module to improve the spatial feature from the spatial-temporal features.  I3D is one kind of the 3DCNN \cite{ji20123d} used to extract 2D or 3D features from dynamic or video to capture both spatial and temporal features from successive frames. In the study, input come from the RGB video, the feature extraction process is expressed as Equation \ref{eq:I3d_feature}:
\begin{equation} \label{eq:I3d_feature}
\mathbf{F}^{I3D} = \text{I3D}(\mathbf{V}_{RGB})
\end{equation}
where \( \mathbf{V}_{RGB} \) denotes the input RGB video frames and \( \mathbf{F}^{I3D} \) represents the feature map generated by the I3D model.
TCA module took the output of I3D feature $\mathbf{F}^{I3D}$  as input to enhance the temporal features \cite{pu2023learning-TCA_PEL}. It mainly uses a self-attention mechanism to incorporate long-range temporal information between frame-level features by extracting strong relationships among the consecutive frames \cite{shao2021temporal_TCA,pu2023learning-TCA_PEL}.
Figure \ref{fig:tca} illustrates the TCA calculation procedure. The output of the I3D module $\mathbf{F}^{I3D}$ we considered here as \(X\), which is projected in the latent space using linear layers to produce the similarity matrix $M$ described in the previous Section under Equation \ref{eq_sim_matrix}, Equation \ref{eq_g_att} and Equation \ref{eq_g_feature_matrix} \cite{shao2021temporal_TCA,pu2023learning-TCA_PEL}. Then, we enhanced the the similarity  matrix using dynamic position encoding (DPE), which is described in Equation \ref{eq:dpe}.The after calculating the Local attention and context features using Equation \ref{eq_l_att}, \ref{eq_l_feature_matrix} and then masekd with similarity matrix \ref{eq_sim_matrix} \cite{shao2021temporal_TCA,pu2023learning-TCA_PEL}.

The final feature \(X^{o}\) is obtained by combining global and local attention heads which described in Equation \ref{eq:final_feature}. After normalizing, we concatenate with a skip connection and use a linear layer to produce the TCA module output:
\begin{equation} \label{eq:tca_feature}
    X^{c} = \text{LN}(X + f_h(\text{Norm}(X^{o})))
\end{equation}
where \(\alpha\) and \(1 - \alpha\) are weights, and \(\text{Norm}(\cdot)\) denotes normalization.

 \Figure[htp](topskip=0pt, botskip=1pt, midskip=0pt)[scale=.30]{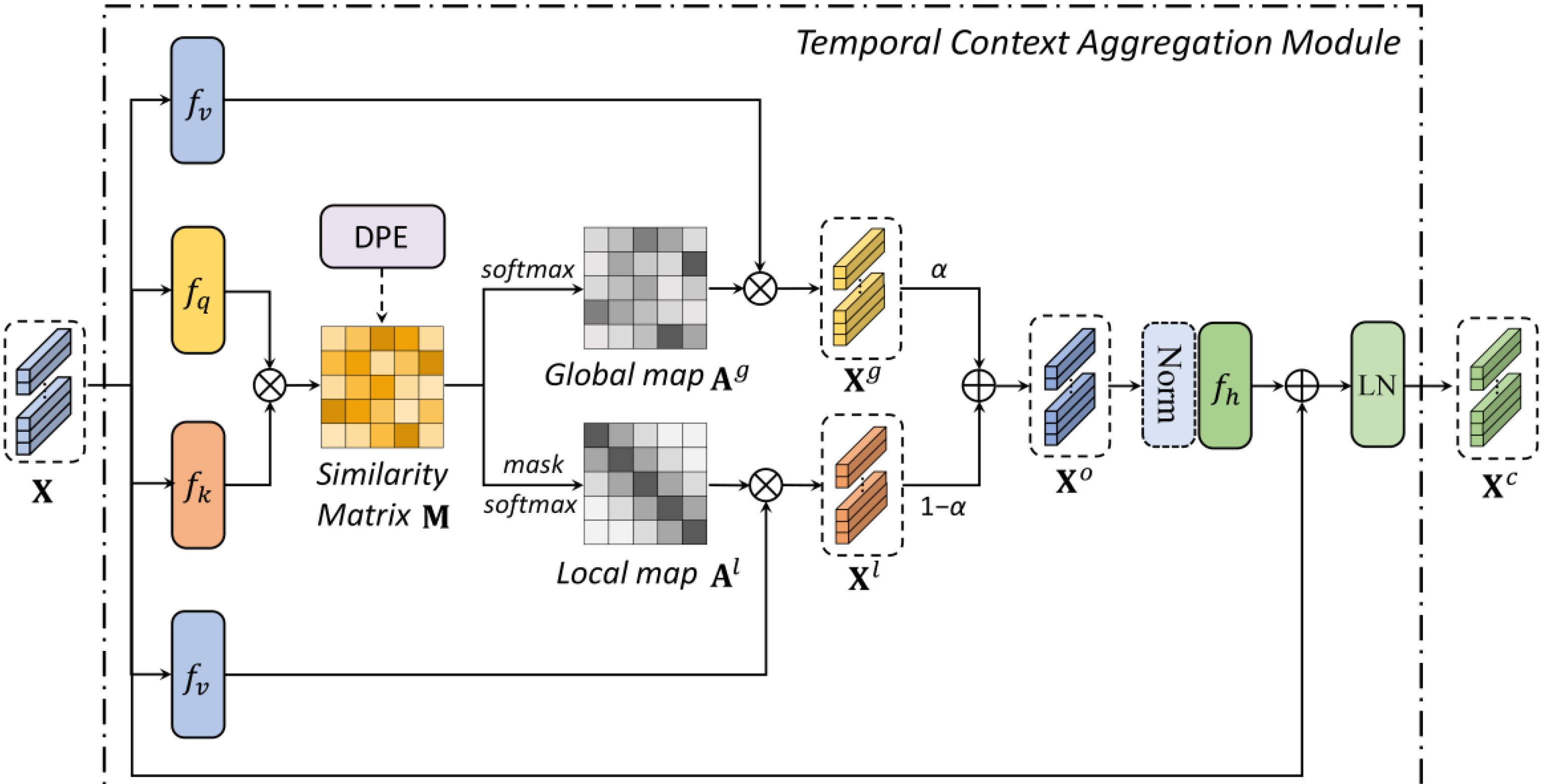}{Working structure of the TCA module \cite{pu2023learning-TCA_PEL,10510436_kaneko_anomaly}\label{fig:tca}}
 Integrating the I3D module with the TCA module mainly extracts robust spatiotemporal contextual information from the video sequence that helps us capture the motion and appearance cues.  TCA plays a pivotal role in integrating contextual information across multiple frames. TCA enhances the model's ability to discern anomalies by considering temporal dependencies within video sequences. This mechanism ensures that the model can effectively capture dynamic changes over time, improving anomaly detection accuracy.  Incorporating a 1D CNN, followed by ReLU activation and dropout regularization, contributes to feature dimensionality reduction while preserving essential information. This process ensures that the extracted features are concise yet informative, facilitating efficient anomaly detection without sacrificing discriminative power.
 \\
\paragraph{Feature Fusion}
In the first stream, we employed the top-k Score Nominator \cite{joo2023cliptsa} to select the top k segments based on their CLIP feature relevance, resulting in a refined set of 512-dimensional features denoted as \(X_{T}\). In the second stream, we obtained the final feature from the FC module, denoted as \(X_{CCN}\). These features were then concatenated, producing comprehensive 1024-dimensional features, denoted as \(F_{stage-1}\), using the following equation:
\begin{equation}
   F_{stage-1}= X_{T} \oplus X_{CCN}
\end{equation}

\subsubsection{Stage 2: UR-DMU Based Feature}
We applied the UR-DMU module to enhance the fused feature, which comes from the attention-based temporal enhancement \cite{URDMU_zh,miah2023dynamic, miah2024spatial_paa,10360810_miah_ksl2,shin2024korean_ksl0}. There are three main components in the UR-DMU shown in Figure \ref{fig:urdmu}. Based on the GCN feature, it construed with Global and Local Multi-Head Self Attention (GL-MHSA) to extract local and global dependency as effective features described in the section \ref{theory:ur-dmu}. For training, videos with normal and abnormal footage are processed. The model generates a score for each snippet using BCE loss and auxiliary losses. During testing, the mean-encoder network of the DUL module produces feature embeddings, which label video snippets to produce UR-DMU features, $F_{urdmu}$. The final feature of stage 2, $X_{Stage-2}$, is obtained by adding $F_{urdmu}$ with TCA $X_{c}$:
\begin{equation} \label{eq:urdmu_tca}
    X_{Stage-2} = F_{urdmu} + X_{c}
\end{equation}

\Figure[htp](topskip=0pt, botskip=1pt, midskip=0pt)[scale=.30]{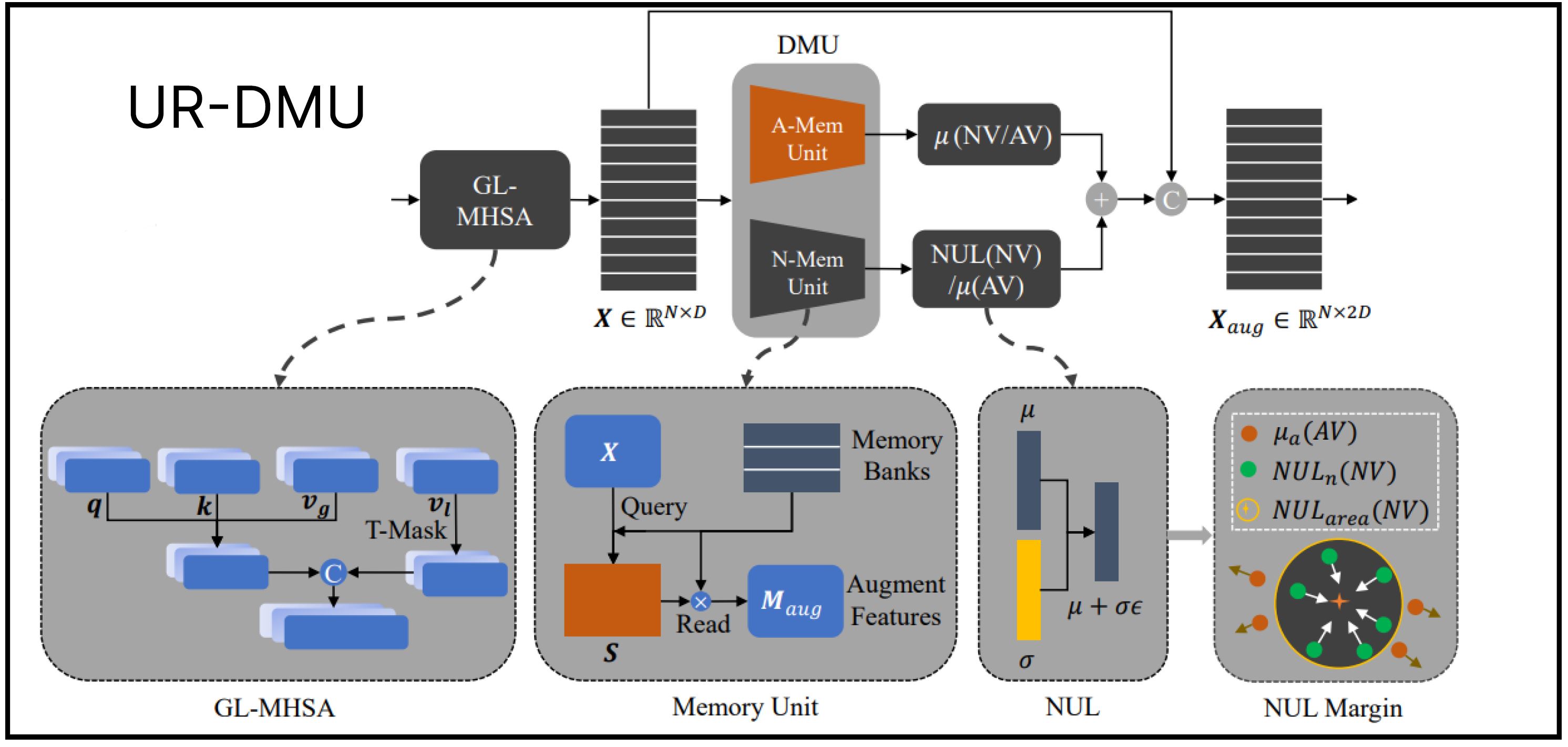}{Working diagram of the UR-DMU module \cite{URDMU_zh,10510436_kaneko_anomaly} \label{fig:urdmu}}

\subsubsection{Stage 3: Feature Reduction MLP Module}
To select effective features from the graph-based UR-DMU $F_{stage-2}$ features, we employed a two-layer MLP for feature reduction. MLP enables high-level semantic representations and non-linear feature transformation for anomaly detection. It includes two Conv1d layers, two GELU activations, and two Dropout mechanisms \cite{zhu2023pdl}. Features from TCA are integrated before the first Conv1d layer, and a 512-dimensional feature from I3D is appended afterwards. Each Conv1D layer is followed by GELU activation and Dropout, as shown in Equation \ref{eq:mlp}:
\begin{equation} \label{eq:mlp}
   \begin{split}
   F_{MLP-1}= \text{Dropout}(\text{GELU}(\text{Conv1D}(F_{stage-2}))) \\
   F_{Stage-3}= \text{Dropout}(\text{GELU}(\text{Conv1D}(F_{MLP-1})))
   \end{split}
\end{equation}
Finally, a causal convolution layer produces anomaly scores by integrating present and past observations, represented as:
\begin{equation} \label{eq:classifier}
    RGB_{Feature} (F)= \sigma\left(f_{t}(X_s)\right),
\end{equation}
where $f_{t}(\cdot)$ denotes the causal convolution layer with a kernel size of $\Delta t$, and $\sigma(\cdot)$ is the sigmoid activation function. The output of the MLP is considered as a final feature denoted as $RGB_{Feature}$
\begin{table*}[t] 
\centering
\caption{Ablation study performance AUC (\%)} \label{ablation}
\begin{tabular}{l|l|l|l|l|l|l|l|l|l|l|l|l|l|l|l}
\textbf{Data Modality} & \textbf{I3D} & \textbf{TCA} & \textbf{CLIP} & \textbf{Top-K} & \textbf{MHSA} & \textbf{DMU} & \textbf{PEL} & \textbf{MC} & \multicolumn{1}{c|}{\textbf{SS}} & \textbf{XD (\%)} & \begin{tabular}[c]{@{}l@{}} \textbf{UCF} \\ \textbf{Crimes} (\%) \end{tabular} & \begin{tabular}[c]{@{}l@{}} \textbf{Shanghai} \\ \textbf{Techs} (\%) \end{tabular} \\ \hline\hline

RGB & \checkmark &  &  &  &  &  &  &  & \multicolumn{1}{c|}{} & 74.82  &  &  \\ \hline
RGB & \checkmark & \checkmark &  &  &  &  &  &  & \multicolumn{1}{c|}{} & 75.63  &  & \\ \hline
RGB & \checkmark & \checkmark & \checkmark &  &  &  &  &  & \multicolumn{1}{c|}{} & 75.63  &  &  \\ \hline
RGB & \checkmark & \checkmark & \checkmark & \checkmark &  &  &  &  & \multicolumn{1}{c|}{} & 76.01  &  &  \\ \hline
RGB & \checkmark & \checkmark & \checkmark & \checkmark & \checkmark &  &  &  & \multicolumn{1}{c|}{} & 79.71  &  &  \\ \hline
RGB & \checkmark & \checkmark & \checkmark & \checkmark & \checkmark & \checkmark &  &  & \multicolumn{1}{c|}{} & 79.74  &  &  \\ \hline
RGB & \checkmark & \checkmark & \checkmark & \checkmark & \checkmark & \checkmark & \checkmark &  & \multicolumn{1}{c|}{} & 83.68  &  &  \\ \hline
RGB & \checkmark & \checkmark & \checkmark & \checkmark & \checkmark & \checkmark & \checkmark & \checkmark & \multicolumn{1}{c|}{} & 86.37  &  &  \\ \hline
RGB & \checkmark & \checkmark & \checkmark & \checkmark & \checkmark & \checkmark & \checkmark & \checkmark & \multicolumn{1}{c|}{\checkmark} & 86.48  & 90.09 & 98.69 \\ \hline\hline
\begin{tabular}[c]{@{}l@{}}RGB+Flow\end{tabular} & \checkmark & \checkmark & \checkmark & \checkmark & \checkmark & \checkmark & \checkmark & \checkmark & \multicolumn{1}{c|}{\checkmark} & 87.32  & 90.26 & 98.71 \\ \hline
\begin{tabular}[c]{@{}l@{}}RGB+Flow+Audio\end{tabular}  & \checkmark & \checkmark & \checkmark & \checkmark & \checkmark & \checkmark & \checkmark & \checkmark & \multicolumn{1}{c|}{\checkmark} & 88.28  &  &  \\ \hline
\end{tabular}

\end{table*}

\subsection{Flow Data Modality Stream}
The Flow Dataset stream took RGB video as input and then was first processed to compute the optical flow, capturing motion dynamics between consecutive frames using \( {\text{TV-L}_1} \)\cite{tvl1}. The optical flow between frames \( t \) and \( t+1 \) is given by:


\begin{equation}
\mathbf{F}_{\text{flow}} = \text{TV-L}_1(\mathbf{V}^{RGB}_t, \mathbf{V}^{RGB}_{t+1})
\label{eq:flow_computation}
\end{equation}

where \( \mathbf{F}_{\text{flow}} \) represents the movement and temporal changes in the video sequence.

\paragraph{I3D Feature Extraction}
After calculating the optical flow, we apply the Inflated 3D ConvNet (I3D) model to extract spatio-temporal features from the flow data. The I3D features are generated as follows:

\begin{equation}
\mathbf{F}_{\text{I3D}} = \text{I3D}(\mathbf{F}_{\text{flow}})
\label{eq:I3D_feature}
\end{equation}

where \( \mathbf{F}_{\text{I3D}} \) captures the complex motion and spatial relationships in the flow information.

\paragraph{MLP Module}
The features extracted by the I3D model are then refined through a Multi-Layer Perceptron (MLP) module to enhance their representation for subsequent processing. The MLP transforms the I3D features as follows:

\begin{equation}
\mathbf{F}_{\text{MLP}} = \text{ReLU}(\mathbf{W}_1 \cdot \mathbf{F}_{\text{I3D}} + \mathbf{b}_1)
\label{eq:MLP_transformation}
\end{equation}

where \( \mathbf{W}_1 \) and \( \mathbf{b}_1 \) are the weights and biases of the MLP, and \( \text{ReLU}(\cdot) \) is the activation function. The output \( \mathbf{F}_{\text{MLP}} \) is a refined representation of the I3D features.

\paragraph{Transformer Module}
The refined features from the MLP are then fed into a Transformer model, which uses self-attention mechanisms to capture complex temporal dependencies within the feature information. The processing by the Transformer is expressed as:

\begin{equation}
\mathbf{F}_{\text{Trans}} = \text{Transformer}(\mathbf{F}_{\text{MLP}})
\label{eq:Transformer_processing}
\end{equation}
where \( \mathbf{F}_{\text{Trans}} \) represents the feature map after Transformer processing.
The final feature representation, \( \mathbf{F}_{\text{Trans}} \), referred to as the "Feature of Flow Modality," encapsulates the enriched temporal dynamics and motion information, ready for downstream tasks such as classification, segmentation, or anomaly detection, providing a comprehensive analysis of the RGB video data. This approach is advantageous because it effectively highlights motion-related anomalies, providing complementary information to the RGB stream. The novelty of this stream lies in the integration of motion flow information with advanced Transformer-based processing, enhancing the model's ability to detect subtle and complex anomalies.

\subsection{Audio Data Modality Stream}
VGGish is a deep convolutional neural network that effectively extracts features from audio signals by treating the spectrogram of the audio signal as an image. Here’s how VGGish is applied for feature extraction from audio signals:
\paragraph{Audio Preprocessing}
The audio signal is first converted into a spectrogram, which represents the frequency content of the signal over time. It is usually computed using the Short-Time Fourier Transform (STFT). The spectrogram \( S \) of an audio signal \( x(t) \) can be defined as:
\begin{equation}
S(t, f) = \left| \sum_{n} x[n] \cdot w[n - t] \cdot e^{-j2\pi fn} \right|
\end{equation}
where \( w[n] \) is a window function, \( t \) represents time, and \( f \) represents frequency.
\paragraph{Spectrogram as Input}
The resulting spectrogram, a 2D representation of the audio signal, is fed into the VGGish network as if it were an image. This allows the network to leverage its pre-trained convolutional layers to extract relevant features.

\paragraph{Feature Extraction with VGGishish}
VGGish consists of 16 convolutional layers followed by 3 fully connected layers. For feature extraction, we typically use the outputs of one of the deeper convolutional layers or the fully connected layers. The convolutional layers perform a series of operations defined by:
\begin{equation}
\text{Conv}(S) = \text{ReLU}(W * S + b)
\end{equation}
Where \( * \) denotes the convolution operation, \( W \) is the filter (kernel), \( b \) is the bias, and ReLU is the activation function. Pooling layers in VGGish help reduce the dimensionality of the feature maps while retaining important information. The max-pooling operation can be expressed as:
\begin{equation}
\text{MaxPool}(S) = \max_{i,j \in R_k} S(i,j)
\end{equation}
where \( R_k \) represents the pooling region.
\paragraph{Integrating VGGish and Transformer}
The high-dimensional feature vector produced by VGGish encapsulates the temporal and spectral characteristics of the audio signal. This feature vector is then fed into the Transformer model to capture complex dependencies within the audio data, leveraging the Transformer's self-attention mechanism. The integration of VGGish and Transformer enhances the ability to model both local and global temporal dependencies, making it particularly effective for anomaly detection tasks.
\paragraph{Audio Modality Feature Representation}
The VGGish model processes the spectrogram through its layers, transforming it into a high-dimensional feature vector. This feature vector, denoted as \( Audio_{Feature}(F) \), is further refined by the Transformer model. The final audio feature representation can be expressed as:
\begin{equation}
Audio_{Feature}(F) = \text{Transformer}(\text{VGGishish}(S))
\end{equation}
By integrating VGGish and Transformer, we effectively transform raw audio data into a rich, high-dimensional feature space suitable for various audio processing tasks, particularly for discerning anomalies.
\subsection{Feature Concatenation and Classification}
To leverage the combined information from different modalities, we perform feature concatenation followed by classification. Specifically, we first concatenate features from the RGB video modality with the Flow modality and then apply a deep learning-based classification module. The process is as follows:
\paragraph{Feature Concatenation}
Let \( F_{rgb} \) represent the feature vector from the RGB video modality and \( F_{flow} \) represent the feature vector from the Flow modality. The concatenated feature vector \( F_{concat} \) is given by:
\begin{equation} \label{eq:concat_rgb_flow}
    F_{concat} = [F_{rgb}; F_{flow}]
\end{equation}
Where \([ ; ]\) denotes concatenation along the feature dimension.
\paragraph{Extended Feature Concatenation}
For scenarios involving additional modalities, such as combining RGB video, Flow, and Audio modalities, let \( F_{audio} \) represent the feature vector from the Audio modality. The extended concatenated feature vector \( F_{extended} \) is given by:

\begin{equation} \label{eq:concat_rgb_flow_audio}
  F_{extended}  = [F_{rgb}; F_{flow}; F_{audio}]
\end{equation}

\subsection{Classification and Procedure of Experiment}
The concatenated feature vector \( F_{concat} \) is then passed through a deep learning-based classification module, which produces the final classification score. Let \( \mathbf{W}_{clf} \) and \( \mathbf{b}_{clf} \) be the weights and biases of the classification layer. The classification score \( S_{concat} \) is computed as:

\begin{equation} \label{eq:classification_rgb_flow}
    S_{concat} = \text{Softmax}(\mathbf{W}_{clf} \cdot F_{concat} + \mathbf{b}_{clf})
\end{equation}
Similarly, this extended feature vector \( F_{extended} \) is used in the classification module to produce performance accuracy. The classification score \( S_{extended} \) is:

\begin{equation} \label{eq:classification_rgb_flow_audio}
    S_{extended} = \text{Softmax}(\mathbf{W}_{clf} \cdot F_{extended} + \mathbf{b}_{clf})
\end{equation}
where Softmax converts the output logits into probabilities.
To evaluate the performance, we use the Area Under the Curve (AUC) metric with Multiple Instance Learning (MIL) and Magniture Contrasive (MC) loss functions.
During training experiment, we optimize the objective function:
\begin{equation} \label{eq:training_objective}
    L = L_{\text{ce}} + \lambda L_{\text{kd}}
\end{equation}
where \( L_{\text{ce}} \) represents the cross-entropy loss, \( L_{\text{kd}} \) denotes the knowledge distillation loss, and \( \lambda \) is a hyperparameter that balances these losses. This formulation enhances the model's ability to differentiate between positive and negative snippets by improving discriminative representations.
During testing, we apply score smoothing (SS) to reduce transient noise and false alarms, as described by:
\begin{equation} \label{eq:ss}
    \tilde{s}_i = \frac{1}{\kappa} \sum_{j=i}^{i+\kappa-1} s_j 
\end{equation}
Here, \( \tilde{s}_i \) represents the smoothed score for the \( i \)-th snippet, and \( \kappa \) is the smoothing window size. This approach suppresses noise and biases, resulting in more stable prediction scores. We do not perform feature-length normalization and treat each video independently. Extracted video features are processed through a Temporal Spatiotemporal Attention Network (TSAN), which generates reweighted attention features. These features are then fed into a snippet association network and an MLP-based converter to produce anomaly scores. Each score, ranging from 0 to 1, reflects the anomaly probability of the corresponding snippet. For evaluation, the anomaly scores are replicated \( \Delta \) times to align with the video's frame length, ensuring accurate assessment across the entire video sequence.

\section{Experimental Evaluation}
We evaluated the proposed model using three anomaly datasets with various modalities. In the section below, we first describe the datasets and then include the performance accuracy for each dataset. 
\subsection{Datasets}
Anomaly detection datasets are essential for developing and testing algorithms that spot unusual events in data streams. XD-Violence \cite{URDMU_zh} is one such dataset that includes a wide range of anomalies in terms of scale, backgrounds, and types. It serves as a valuable resource for researchers to train and evaluate their anomaly detection models across diverse scenarios. By using these datasets, researchers can benchmark their methods, evaluate performance, and contribute to enhancing anomaly detection applications in real-world settings. In the study, we used three modalities datasets, including RGB Video, Flow and Audio signal information; we found only one dataset which contained three modalities. In addition, we also used the ShanghaiTech dataset and UCF-Crime dataset, which consisted of only RGB and Flow information. 
\paragraph{XD-Violence}
 The XD-Violence dataset contains a mix of video and audio media formats, covering diverse backgrounds like movies, games, and live scenes. It includes 4,754 videos in total, with 3,954 videos for training, each labelled at the video level. Additionally, there are 800 testing videos labelled frame by frame \cite{URDMU_zh}.
\paragraph{Other Datasets}
We also used the ShanghaiTech and UCF-Crime datasets, which include RGB and Flow data but no audio. The ShanghaiTech dataset contains 317,398 frames from various locations on the ShanghaiTech Campus, with 307 normal and 130 anomaly videos across 13 scenes. Originally a benchmark for video anomaly detection, the dataset was reorganized by Zhong et al. to create a weakly supervised training set. We followed their approach for our experiments \cite{zhong2019graph}.UCF-Crime Dataset. The UCF-Crime dataset consists of 1,900 untrimmed videos, totalling 128 hours, featuring 13 types of real-world anomalies like arson, burglary, and robbery. It offers more complex backgrounds than ShanghaiTech. The training set has 1,610 videos (800 normal, 810 anomalous), while the testing set includes 290 videos with frame-level labels \cite{10510436_kaneko_anomaly}.

\subsubsection{Environmental Setup and Evaluation Metrices}
The system was built with a GeForce RTX 4090 24GB GPU, CUDA version 11.7, NVIDIA driver 515, and 64GB of RAM. The training utilized two learning rates of 0.00003 for the flow data flow and the audio data flow and 0.000001 for the rgb data flow, and batch size 32, and ran for two epochs using the Adam optimizer on the RTX 4090. For efficient graph convolution and attention with low computational cost, the Python environment included OpenCV, Pickle, Pandas, and PyTorch for model development \cite{paszke2019pytorch}. These packages, along with others \cite{gollapudi2019learn, dozat2016incorporating}, facilitated initial data processing and model development.

\begin{table}[t] 
\centering
\setlength{\tabcolsep}{10pt}
\caption{Performance result} \label{tab:peformance_result}
  \centering
  \begin{adjustwidth}{0cm}{0cm}
\centering
   \begin{tabular}{lllll}
     \hline
     \begin{tabular}[c]{@{}l@{}}Dataset Name   \end{tabular}  & AUC (\%)  &\begin{tabular}[c]{@{}l@{}}Anomaly\\ AUC (\%)   \end{tabular}  & AP (\%) & FAR (\%) \\ \hline
     \begin{tabular}[c]{@{}l@{}}XD-Violence   \end{tabular}  &95.84&86.92 & 88.28& 0.0014   \\ \hline
     \end{tabular}
\end{adjustwidth}
\end{table}

\begin{table}[htbp]
\centering
\caption{State-of-the-art comparison of the proposed model for the XD Violence Dataset}
\label{tab:sota_xdviolence_dataset}
\begin{tabular}{lll}
\hline
\textbf{Method}        & \textbf{Feature} & \textbf{AP (\%)} \\ \hline
Sultani et al.\cite{sultani2018real}  & RGB              & 73.20            \\ 
HL-Net \cite{wu2020not}          & RGB              & 73.67            \\ 

RTFM \cite{RTFM_tian2021weakly}           & RGB              & 77.81            \\ 
MSL \cite{li2022self}             & RGB              & 78.28            \\ 
MSL \cite{li2022self}             & RGB              & 78.59            \\ 
HL-Net \cite{wu2020not}          & RGB+Audio        & 78.64            \\ 
Pang et al. \cite{pang2020self}     & RGB+Audio        & 81.69            \\ 
ACF \cite{ACF_wei2022look}             & RGB+Audio        & 80.13            \\
MSAF \cite{9926192}          & RGB+Audio          & 80.51            \\
CUPL \cite{zhang2023exploiting}          & RGB+Audio          & 81.43            \\
CMA-LA \cite{9712793}          & RGB+Audio          &  83.54            \\
MACIL-SD \cite{yu2022modality}          & RGB+Audio          &  83.40            \\
Pu et al. \cite{pu2023learning-TCA_PEL}          & RGB          & 85.59            \\ \hline
Shin et al. \cite{10510436_kaneko_anomaly}          & RGB        & 86.26            \\ \hline
Propsoed Multimodal       & RGB+Flow+Audio        & 88.28            \\ \hline
\end{tabular}
\end{table}

\begin{table*}[t] 
\caption{State-of-the-art comparison of the proposed model for the UCF Crime and ShanghaiTech datasets}
\centering
\label{tab:sota_ucf_shanghai_dataset}
\setlength{\tabcolsep}{10pt} 
\begin{adjustwidth}{0cm}{0cm}
\begin{tabular}{lllllllll}
\hline
\multirow{2}{*}{Model Name} & \multicolumn{2}{c}{Data Modalities} & \multirow{2}{*}{Year} & \multirow{2}{*}{Feature Extractor Name} & \multicolumn{2}{l}{ShanghaiTech Dataset} & \multicolumn{2}{l}{UCF Crime Dataset} \\ 
& Video & Flow  &   &   & AUC(\%)  & & AUC(\%)  \\ \hline
Sultani et al. \cite{sultani2018real}& \checkmark & - & 2018  &  C3D & 83.17 &   & 75.41  \\
Sultani et al. \cite{sultani2018real} & \checkmark & - & 2018  &  ID3 & 85.33 &  & 77.92  \\
\hline
Zhong et al. \cite{zhong2019graph}& \checkmark & - & 2019  &  C3D & 76.44 &   & 81.08 \\
Zhong et al. \cite{zhong2019graph}& \checkmark & - & 2019  &  TSN & 84.44 &   & 82.12  \\
Zhong et al. \cite{zhang2019temporal} & \checkmark & - & 2019  &  ID3 & 82.50 &  & 78.70 \\
\hline
Zaheer et al. \cite{zaheer2020self}& \checkmark & - & 2020  &  C3D-self & 84.16 &   & 79.54 \\
Zaheer et al. \cite{zaheer2020claws}& \checkmark & - & 2020  &  C3D & 89.67 &   & 83.03  \\
Wan et al. \cite{wan2020weakly} & \checkmark & - & 2020  & I3D & 85.38 &  & 78.96 \\
\hline
Purwanto et al. \cite{purwanto2021dance}& \checkmark & - & 2021  &  TRN & 96.85 & & 85.00 \\
Tian et al. \cite{RTFM_tian2021weakly}& \checkmark & - & 2021  &  C3D & 91.51 & & 83.28 \\
Majhietal et al. \cite{majhi2021dam} & \checkmark & - & 2021  & ID3 & 88.22 & & 82.67 \\
Tianetal et al. \cite{RTFM_tian2021weakly} & \checkmark & - & 2021  & ID3 & 97.21 & & 84.30 \\
Wuetal et al. \cite{wu2021learning_mil} & \checkmark & - & 2021  & ID3 & 97.48 & & 84.89 \\
Yuetal et al. \cite{yu2021cross} & \checkmark & - & 2021  & ID3 & 87.83 & & 82.15 \\
Lvetal et al. \cite{lv2021localizing} & \checkmark & - & 2021  & ID3 & 85.30 & & 85.38 \\
Fengetal et al. \cite{feng2021mist} & \checkmark & - & 2021  & CD3 & 93.13 & & 81.40 \\
\hline
Zaheer et al. \cite{zaheer2022generative} & \checkmark & - & 2022  & ResNext & 86.21 & & 79.84 \\
Zaheer et al. \cite{zaheer2023clustering} & \checkmark & - & 2022  & CD3 & 90.12 & & 83.37 \\
Zaheer et al. \cite{zaheer2023clustering} & \checkmark & - & 2022  & 3DResNext & 91.46 & & 84.16 \\
Joo et al.\cite{joo2212clip} & \checkmark & - & 2022  & C3D & 97.19 & & 83.94 \\
Joo et al. et al. \cite{joo2212clip} & \checkmark & - & 2022  & I3D & 97.98 & & 84.66 \\
Joo et al. et al. \cite{joo2212clip} & \checkmark & - & 2022  & CLIP & 98.32 & & 87.58 \\
Cao et al. \cite{cao2023weakly} & \checkmark & - & 2022  & I3D & 96.45 & & 85.87 \\
Li et al. \cite{li2022self} & \checkmark & - & 2022  & I3D & 96.08 & & 85.30 \\
Cao et al. \cite{cao2023weakly} & \checkmark & - & 2022  & I3D-graph & 96.05 & & 84.67 \\
Tan et al. \cite{tan2024overlooked} & \checkmark & - & 2022  & I3D & 97.54 & & 86.71 \\
Li et al. \cite{li2022self} & \checkmark & - & 2022  & VideoSwim & 97.32 & & 85.62 \\
Yi et al. \cite{yi2022batch} & \checkmark & - & 2022  & I3D & 97.65 & & 84.29 \\
Yu et al. \cite{yu2022tca} & \checkmark & - & 2022  & C3D & 88.35 & & 82.08 \\
Yu et al. \cite{yu2022tca} & \checkmark & - & 2022  & I3D & 89.91 & & 83.75 \\
Gong et al. \cite{gong2022multi} & \checkmark & - & 2022  & I3D & 90.10 & & 81.00 \\
\hline
Majhi et al. \cite{majhi2023human} & \checkmark & - & 2023  & 13D-Res & 96.22 & & 85.30 \\
Park et al. \cite{park2023normality} & \checkmark & - & 2023  & C3D & 96.02 & & 83.43 \\
Park et al. \cite{park2023normality} & \checkmark & - & 2023  & I3D & 97.43 & & 85.63 \\
Pu  et al. \cite{pu2023learning-TCA_PEL} & \checkmark & - & 2023 & I3D & 98.14 & & 86.76 \\
Lv et al. \cite{lv2023unbiased} & \checkmark & - & 2023  & X-CLIP & 96.78 & & 86.75 \\
Sun et al. \cite{sun2023long} & \checkmark & - & 2023  & C3D & 96.56 & & 83.47 \\
Sun et al. \cite{sun2023long} & \checkmark & - & 2023  & I3D & 97.92 & & 85.88 \\
Wang et al. \cite{wang2023attention} & \checkmark & - & 2023  & C3D & 94.01 & & 81.48 \\
Sharif et al. \cite{sharif2023cnn_CNN-ViT} & \checkmark & - & 2023  & I3D+CLIP & 98.66 & & 88.97 \\
Shin et al. \cite{10510436_kaneko_anomaly}  & \checkmark & - & 2024  & hybrid model & 98.69 & & 90.00 \\ 
Proposed Model & \checkmark & \checkmark & -  & Multimodality & 98.71 & & 90.26 \\ 
\hline
\end{tabular}
\end{adjustwidth}
\end{table*}

\subsection{Ablation Study}
Table \ref{ablation} presents the ablation study results for the proposed model, highlighting the contributions of various components, including the multi-backbone pre-trained models. In this study, we systematically evaluated the impact of different technologies on weakly supervised video anomaly detection. The check marks indicate the utilization of the corresponding technology in our experiments.
Starting with the I3D module alone, performance significantly improved, achieving an AUC of 74.82\% on the XD dataset. When the Temporal Contextual Aggregation (TCA) module was added, performance further improved to 75.63\%, demonstrating the benefit of capturing temporal dependencies. The integration of the CLIP module with TCA and I3D maintains the AUC at 75.63\%, but adding the Top-K selection mechanism boosts it to 76.01\%, indicating the importance of selecting the most relevant features. The introduction of Multi-Head Self Attention (MHSA) leads to a substantial increase in AUC to 79.71\%, highlighting the role of attention mechanisms in focusing on critical information.
Further enhancement with the Dual Memory Unit (DMU) slightly increases the AUC to 79.74\%. The incorporation of Memory Combination (MC) further increases the AUC to 86.37\%. Finally, applying Score Smoothing (SS) alongside all aforementioned technologies results in the highest detection accuracy, achieving an AUC of 86.48\%. Next, we extended the model to include the RGB modality, achieving a score of 86.37\%. We then combined the flow modality with a Gated Feature Fusion with Attention Module before the final layer. The weights of the RGB modality component were initialized using the model that scored 86.37\%, allowing us to reach high accuracy without extensive retraining. As a result, we achieved 87.32\% on the XD-Violence dataset, demonstrating the effectiveness of the Gated Feature Fusion with Attention Module and the inclusion of the flow modality.
Similarly, we developed a model that incorporates the audio modality in addition to the flow modality, with the weights initialized only for the RGB portion, as was done previously with the flow modality. This approach resulted in a very high accuracy of 88.32\%, highlighting the usefulness of the audio modality.
This demonstrates the synergistic effect of combining these methodologies, showcasing the robustness and effectiveness of the multimodal approach in improving video anomaly detection performance.

\subsection{Performance Result and State of the Art Comparison for XD Violence Dataset}
Table \ref{tab:peformance_result} demonstrates the performance of the proposed model. The proposed model demonstrates high performance on the XD-Violence dataset with an AUC of 95.84\%, an anomaly AUC of 86.92\%, an AP of 88.28\%, and a FAR of 0.0014\%. These results highlight the model's accuracy and reliability in detecting anomalies.
Table \ref{tab:sota_xdviolence_dataset} demonstrated the performance and state of the part comparison of the proposed model for video anomaly detection on the XD Violence Dataset, highlighting the performance of the proposed multimodal model. Sultani et al. \cite{sultani2018real} achieved an average precision (AP) of 73.20\% using RGB features. HL-Net \cite{wu2020not} improved performance to 73.67\% with RGB and further to 78.64\% by incorporating audio. RTFM \cite{RTFM_tian2021weakly} reached 77.81\% using RGB features. MSL \cite{li2022self} showed a significant enhancement, achieving up to 78.59\% with RGB features.
Pang et al. \cite{pang2020self} reported an AP of 81.69\% using RGB and audio, while ACF \cite{ACF_wei2022look} and MSAF \cite{9926192} achieved 80.13\% and 80.51\% respectively with the same feature set. CUPL \cite{zhang2023exploiting} further improved to 81.43\%. CMA-LA \cite{9712793} and MACIL-SD \cite{yu2022modality} achieved high APs of 83.54\% and 83.40\%, respectively, with RGB and audio features. Pu et al. \cite{pu2023learning-TCA_PEL} achieved an impressive 85.59\% using RGB alone, while Shin et al. \cite{10510436_kaneko_anomaly} reached 86.26\%.
The proposed multimodal model, integrating RGB, flow, and audio features, surpasses all previous methods with an AP of 88.28\%, demonstrating the effectiveness of combining multiple data modalities for enhanced anomaly detection performance.

\subsection{State of the Art Comparison for Other Datasets}
We also evaluated the proposed model using the ShanghaiTech Dataset and UCF Crime Dataset. Our model is designed to incorporate three modality features: RGB Video, Flow, and Audio information. However, due to the limited availability of audio modality datasets, we focused our evaluation on the RGB Video and Flow modalities. The table in Figure \ref{tab:sota_ucf_shanghai_dataset} presents a state-of-the-art comparison of these datasets, demonstrating the effectiveness of our approach. The proposed multimodality model achieved an AUC of 98.71\% on the ShanghaiTech dataset and 90.26\% on the UCF Crime dataset, outperforming existing state-of-the-art models. This strong performance highlights the significant advantage of integrating multiple data modalities. By leveraging both RGB and Flow information, our model can capture more complex patterns and nuances in the data, leading to more accurate anomaly detection. This approach is particularly effective in diverse and dynamic environments, where relying on a single modality might result in missed or inaccurate detections. Our multimodal model's superior accuracy and robustness make it a strong candidate for real-world applications in surveillance and security, offering a more comprehensive solution than models that rely on single data types.

\section{Conclusion }
This study presented a comprehensive multimodal deep learning model for weakly supervised video anomaly detection (WS-VAD), leveraging RGB video, optical flow, and audio data modalities. Our model integrates advanced feature extraction techniques, including a ViT-based CLIP module, CNN-based I3D with Temporal Context Aggregation (TCA), and Uncertainty-Resilient Dual Memory Units (UR-DMU) with Global/Local Multi-Head Self Attention (GL-MHSA) and Transformer. These components, coupled with a multilayer perceptron (MLP) for feature refinement, significantly enhance the model's ability to distinguish between normal and abnormal behaviours. Each modality contributes uniquely: the RGB stream captures visual semantics, the flow stream emphasizes dynamic motion, and the audio stream detects anomalies through sound patterns. Integrating these streams via a gated feature fusion mechanism with an attention module creates a robust classifier that effectively predicts snippet-level anomaly scores and converts them into bag-level predictions during training. Extensive experiments on three benchmark datasets demonstrate that our model surpasses existing state-of-the-art approaches, delivering high accuracy and robust performance. This model shows great promise for real-world applications, offering a reliable and efficient solution for intelligent surveillance systems.

\section*{ABBREVIATIONS}
\begin{table}[H]
\label{Appendix_A}
\setlength{\tabcolsep}{3pt}
\begin{tabular}{ll}
WS-VAD & weakly supervised video anomaly detection \\
TCA & Temporal Contextual Aggregation\\ 
GL-MHSA &  Global/Local Multi-Head Self Attention\\
CNN& convolutional network\\
ViT& vision transformer\\
UR-DMU & Uncertainty-regulated Dual Memory Units \\
MLP & Multilayer Perceptron \\
 MIL & multiple instances
learning \\ 
NVs&Normal videos\\
AVs &Anomalous videos\\
DL& Deep learning\\
BCE&  Binary Cross-Entropy\\
\end{tabular} 
\end{table}

\newpage
\bibliographystyle{unsrt}
\bibliography{sn-bibliography}

\begin{thebibliography}{10}

\bibitem{liu2019exploring}
Kun Liu and Huadong Ma.
\newblock Exploring background-bias for anomaly detection in surveillance videos.
\newblock In {\em Proceedings of the 27th ACM International Conference on Multimedia}, pages 1490--1499, Nice, France, 2019.

\bibitem{sharif2023deep}
Md~Haidar Sharif, Lei Jiao, and Christian~W Omlin.
\newblock Deep crowd anomaly detection by fusing reconstruction and prediction networks.
\newblock {\em Electronics}, 12(7):1517, 2023.

\bibitem{chandola2009anomaly}
Varun Chandola, Arindam Banerjee, and Vipin Kumar.
\newblock Anomaly detection: A survey.
\newblock {\em ACM computing surveys (CSUR)}, 41(3):1--58, 2009.

\bibitem{zhong2019graph}
Jia-Xing Zhong, Nannan Li, Weijie Kong, Shan Liu, Thomas~H Li, and Ge~Li.
\newblock Graph convolutional label noise cleaner: Train a plug-and-play action classifier for anomaly detection.
\newblock In {\em Proceedings of the IEEE/CVF conference on computer vision and pattern recognition}, pages 1237--1246, Long Beach, CA, USA, 2019.

\bibitem{zaheer2020claws}
Muhammad~Zaigham Zaheer, Arif Mahmood, Marcella Astrid, and Seung-Ik Lee.
\newblock Claws: Clustering assisted weakly supervised learning with normalcy suppression for anomalous event detection.
\newblock In {\em Computer Vision--ECCV 2020: 16th European Conference, Glasgow, UK, August 23--28, 2020, Proceedings, Part XXII 16}, pages 358--376. Springer, 2020.

\bibitem{sultani2018real}
Waqas Sultani, Chen Chen, and Mubarak Shah.
\newblock Real-world anomaly detection in surveillance videos.
\newblock In {\em Proceedings of the IEEE conference on computer vision and pattern recognition}, pages 6479--6488, Salt Lake City, UT, USA, 2018.

\bibitem{zhang2019temporal}
Jiangong Zhang, Laiyun Qing, and Jun Miao.
\newblock Temporal convolutional network with complementary inner bag loss for weakly supervised anomaly detection.
\newblock In {\em 2019 IEEE International Conference on Image Processing (ICIP)}, pages 4030--4034, Taipei, Taiwan, 2019. IEEE.

\bibitem{ullah2021weakly}
Sami Ullah, Naeem Bhatti, Tehreem Qasim, Najmul Hassan, and Muhammad Zia.
\newblock Weakly-supervised action localization based on seed superpixels.
\newblock {\em Multimedia Tools and Applications}, 80:6203--6220, 2021.

\bibitem{zhu2019motion}
Yi~Zhu and Shawn Newsam.
\newblock Motion-aware feature for improved video anomaly detection.
\newblock {\em arXiv preprint arXiv:1907.10211}, 2019.

\bibitem{lv2021localizing}
Hui Lv, Chuanwei Zhou, Zhen Cui, Chunyan Xu, Yong Li, and Jian Yang.
\newblock Localizing anomalies from weakly-labeled videos.
\newblock {\em IEEE transactions on image processing}, 30:4505--4515, 2021.

\bibitem{hassan2018temporal}
Najmul Hassan, Naeem Bhatti, et~al.
\newblock Temporal superpixels based human action localization.
\newblock In {\em 2018 14th International Conference on Emerging Technologies (ICET)}, pages 1--6. IEEE, 2018.

\bibitem{purwanto2021dance}
Didik Purwanto, Yie-Tarng Chen, and Wen-Hsien Fang.
\newblock Dance with self-attention: A new look of conditional random fields on anomaly detection in videos.
\newblock In {\em Proceedings of the IEEE/CVF International Conference on Computer Vision}, pages 173--183, Montreal, BC, Canada, 2021.

\bibitem{miah2023dynamic}
Abu Saleh~Musa Miah, Md~Al~Mehedi Hasan, and Jungpil Shin.
\newblock Dynamic hand gesture recognition using multi-branch attention based graph and general deep learning model.
\newblock {\em IEEE Access}, 11:4703--4716, 2023.

\bibitem{miah2024spatial_paa}
Abu Saleh~Musa Miah, Md~Al~Mehedi Hasan, Yuichi Okuyama, Yoichi Tomioka, and Jungpil Shin.
\newblock Spatial-temporal attention with graph and general neural network-based sign language recognition.
\newblock {\em Pattern Analysis and Applications}, 27(2):37, 2024.

\bibitem{miah2024_multiculture}
Abu Saleh~Musa Miah, Md~Al~Mehedi Hasan, Yoichi Tomioka, and Jungpil Shin.
\newblock Hand gesture recognition for multi-culture sign language using graph and general deep learning network.
\newblock {\em IEEE Open Journal of the Computer Society}, 2024.

\bibitem{carbonneau2018multiple}
Marc-Andr{\'e} Carbonneau, Veronika Cheplygina, Eric Granger, and Ghyslain Gagnon.
\newblock Multiple instance learning: A survey of problem characteristics and applications.
\newblock {\em Pattern Recognition}, 77:329--353, 2018.

\bibitem{liu2023generalized}
Yang Liu, Dingkang Yang, Yan Wang, Jing Liu, and Liang Song.
\newblock Generalized video anomaly event detection: Systematic taxonomy and comparison of deep models.
\newblock {\em arXiv preprint arXiv:2302.05087}, 2023.

\bibitem{wu2020not}
Peng Wu, Jing Liu, Yujia Shi, Yujia Sun, Fangtao Shao, Zhaoyang Wu, and Zhiwei Yang.
\newblock Not only look, but also listen: Learning multimodal violence detection under weak supervision.
\newblock In {\em Computer Vision--ECCV 2020: 16th European Conference, Glasgow, UK, August 23--28, 2020, Proceedings, Part XXX 16}, pages 322--339. Springer, 2020.

\bibitem{RTFM_tian2021weakly}
Yu~Tian, Guansong Pang, Yuanhong Chen, Rajvinder Singh, Johan~W Verjans, and Gustavo Carneiro.
\newblock Weakly-supervised video anomaly detection with robust temporal feature magnitude learning.
\newblock In {\em Proceedings of the IEEE/CVF international conference on computer vision}, pages 4975--4986, Montreal, BC, Canada, 2021.

\bibitem{joo2212clip}
HK~Joo, K~Vo, K~Yamazaki, and N~Le.
\newblock Clip-tsa: Clip-assisted temporal self-attention for weakly-supervised video anomaly detection. arxiv 2022.
\newblock {\em arXiv preprint arXiv:2212.05136}.

\bibitem{ji20123d}
Shuiwang Ji, Wei Xu, Ming Yang, and Kai Yu.
\newblock 3d convolutional neural networks for human action recognition.
\newblock {\em IEEE transactions on pattern analysis and machine intelligence}, 35(1):221--231, 2012.

\bibitem{carreira2017quo}
Joao Carreira and Andrew Zisserman.
\newblock Quo vadis, action recognition? a new model and the kinetics dataset.
\newblock In {\em proceedings of the IEEE Conference on Computer Vision and Pattern Recognition}, pages 6299--6308, Honolulu, HI, USA, 2017.

\bibitem{shao2021temporal_TCA}
Jie Shao, Xin Wen, Bingchen Zhao, and Xiangyang Xue.
\newblock Temporal context aggregation for video retrieval with contrastive learning.
\newblock In {\em Proceedings of the IEEE/CVF winter conference on applications of computer vision}, pages 3268--3278, Virtual Conference, 2021.

\bibitem{yu2022tca}
Shenghao Yu, Chong Wang, Lehong Xiang, and Jiafei Wu.
\newblock Tca-vad: Temporal context alignment network for weakly supervised video anomly detection.
\newblock In {\em 2022 IEEE International Conference on Multimedia and Expo (ICME)}, pages 1--6, Taipei, Taiwan, 2022. IEEE.

\bibitem{pu2023learning-TCA_PEL}
Yujiang Pu, Xiaoyu Wu, and Shengjin Wang.
\newblock Learning prompt-enhanced context features for weakly-supervised video anomaly detection.
\newblock {\em arXiv preprint arXiv:2306.14451}, 2023.

\bibitem{URDMU_zh}
Hang Zhou, Junqing Yu, and Wei Yang.
\newblock Dual memory units with uncertainty regulation for weakly supervised video anomaly detection.
\newblock In {\em Proceedings of the AAAI Conference on Artificial Intelligence}, volume~37, pages 3769--3777, Washington, DC, USA, 2023.

\bibitem{sharif2023cnn_CNN-ViT}
Md~Haidar Sharif, Lei Jiao, and Christian~W Omlin.
\newblock Cnn-vit supported weakly-supervised video segment level anomaly detection.
\newblock {\em Sensors}, 23(18):7734, 2023.

\bibitem{pang2020self}
Guansong Pang, Cheng Yan, Chunhua Shen, Anton van~den Hengel, and Xiao Bai.
\newblock Self-trained deep ordinal regression for end-to-end video anomaly detection.
\newblock In {\em Proceedings of the IEEE/CVF conference on computer vision and pattern recognition}, pages 12173--12182, Seattle, WA, USA, 2020.

\bibitem{ACF_wei2022look}
Dong-Lai Wei, Chen-Geng Liu, Yang Liu, Jing Liu, Xiao-Guang Zhu, and Xin-Hua Zeng.
\newblock Look, listen and pay more attention: Fusing multi-modal information for video violence detection.
\newblock In {\em ICASSP 2022-2022 IEEE International Conference on Acoustics, Speech and Signal Processing (ICASSP)}, pages 1980--1984, Singapore, 2022. IEEE.

\bibitem{zhang2023exploiting}
Chen Zhang, Guorong Li, Yuankai Qi, Shuhui Wang, Laiyun Qing, Qingming Huang, and Ming-Hsuan Yang.
\newblock Exploiting completeness and uncertainty of pseudo labels for weakly supervised video anomaly detection.
\newblock In {\em Proceedings of the IEEE/CVF Conference on Computer Vision and Pattern Recognition}, pages 16271--16280, Vancouver, BC, Canada, 2023.

\bibitem{9712793}
Yujiang Pu and Xiaoyu Wu.
\newblock Audio-guided attention network for weakly supervised violence detection.
\newblock In {\em 2022 2nd International Conference on Consumer Electronics and Computer Engineering (ICCECE)}, pages 219--223, Guangzhou, China, 2022.

\bibitem{yu2022modality}
Jiashuo Yu, Jinyu Liu, Ying Cheng, Rui Feng, and Yuejie Zhang.
\newblock Modality-aware contrastive instance learning with self-distillation for weakly-supervised audio-visual violence detection.
\newblock In {\em Proceedings of the 30th ACM International Conference on Multimedia}, pages 6278--6287, Lisboa, Portugal, 2022.

\bibitem{9926192}
Donglai Wei, Yang Liu, Xiaoguang Zhu, Jing Liu, and Xinhua Zeng.
\newblock Msaf: Multimodal supervise-attention enhanced fusion for video anomaly detection.
\newblock {\em IEEE Signal Processing Letters}, 29:2178--2182, 2022.

\bibitem{hassan2024deep_har_miah}
Najmul Hassan, Abu Saleh~Musa Miah, and Jungpil Shin.
\newblock A deep bidirectional lstm model enhanced by transfer-learning-based feature extraction for dynamic human activity recognition.
\newblock {\em Applied Sciences}, 14(2):603, 2024.

\bibitem{10624624_lstm}
Najmul Hassan, Abu Saleh~Musa Miah, and Jungpil Shin.
\newblock Enhancing human action recognition in videos through dense-level features extraction and optimized long short-term memory.
\newblock In {\em 2024 7th International Conference on Electronics, Communications, and Control Engineering (ICECC)}, pages 19--23, 2024.

\bibitem{mallik2024virtual}
Bijon Mallik, Md~Abdur Rahim, Abu Saleh~Musa Miah, Keun~Soo Yun, and Jungpil Shin.
\newblock Virtual keyboard: A real-time hand gesture recognition-based character input system using lstm and mediapipe holistic.
\newblock {\em Comput. Syst. Sci. Eng.}, 48(2):555--570, 2024.

\bibitem{patashnik2021styleclip}
Or~Patashnik, Zongze Wu, Eli Shechtman, Daniel Cohen-Or, and Dani Lischinski.
\newblock Styleclip: Text-driven manipulation of stylegan imagery.
\newblock In {\em Proceedings of the IEEE/CVF International Conference on Computer Vision}, pages 2085--2094, Montreal, BC, Canada, 2021.

\bibitem{vo2023aoe}
Khoa Vo, Sang Truong, Kashu Yamazaki, Bhiksha Raj, Minh-Triet Tran, and Ngan Le.
\newblock Aoe-net: Entities interactions modeling with adaptive attention mechanism for temporal action proposals generation.
\newblock {\em International Journal of Computer Vision}, 131(1):302--323, 2023.

\bibitem{kashu2022vltint}
Khoa~Vo Kashu~Yamazaki, Sang Truong, Bhiksha Raj, and Ngan Le.
\newblock Vltint: Visual-linguistic transformer-in-transformer for coherent video paragraph captioning.
\newblock {\em arXiv preprint arXiv:2211.15103}, 2022.

\bibitem{Farhan_attention_miah}
Sumaiya Tabssum Mou et~al. Farhanul~Haque, Md. Al-Hasan.
\newblock Multichannel attention networks with ensembledtransfer learning to recognize bangla handwrittencharecte.
\newblock {\em arXiv preprint arXiv: arXiv:2408.10955}, 2024.

\bibitem{10510436_kaneko_anomaly}
Jungpil Shin, Yuta Kaneko, Abu Saleh~Musa Miah, Najmul Hassan, and Satoshi Nishimura.
\newblock Anomaly detection in weakly supervised videos using multistage graphs and general deep learning based spatial-temporal feature enhancement.
\newblock {\em IEEE Access}, 12:65213--65227, 2024.

\bibitem{tran2015learning}
Du~Tran, Lubomir Bourdev, Rob Fergus, Lorenzo Torresani, and Manohar Paluri.
\newblock Learning spatiotemporal features with 3d convolutional networks.
\newblock In {\em Proceedings of the IEEE international conference on computer vision}, pages 4489--4497, Santiago, Chile, 2015.

\bibitem{miah2022bensignnet}
Abu Saleh~Musa Miah, Jungpil Shin, Md~Al~Mehedi Hasan, and Md~Abdur Rahim.
\newblock Bensignnet: Bengali sign language alphabet recognition using concatenated segmentation and convolutional neural network.
\newblock {\em Applied Sciences}, 12(8):3933, 2022.

\bibitem{computers12010013_multistage_musa}
Abu Saleh~Musa Miah, Md. Al~Mehedi Hasan, Jungpil Shin, Yuichi Okuyama, and Yoichi Tomioka.
\newblock Multistage spatial attention-based neural network for hand gesture recognition.
\newblock {\em Computers}, 12(1), 2023.

\bibitem{electronics12132841_multistream}
Abu Saleh~Musa Miah, Md. Al~Mehedi Hasan, Si-Woong Jang, Hyoun-Sup Lee, and Jungpil Shin.
\newblock Multi-stream general and graph-based deep neural networks for skeleton-based sign language recognition.
\newblock {\em Electronics}, 12(13), 2023.

\bibitem{shin2024korean_ksl0}
Jungpil Shin, Abu Saleh~Musa Miah, Yuto Akiba, Koki Hirooka, Najmul Hassan, and Yong~Seok Hwang.
\newblock Korean sign language alphabet recognition through the integration of handcrafted and deep learning-based two-stream feature extraction approach.
\newblock {\em IEEE Access}, 2024.

\bibitem{shin2023korean_ksl1}
Jungpil Shin, Abu~Saleh Musa~Miah, Md~Al~Mehedi Hasan, Koki Hirooka, Kota Suzuki, Hyoun-Sup Lee, and Si-Woong Jang.
\newblock Korean sign language recognition using transformer-based deep neural network.
\newblock {\em Applied Sciences}, 13(5):3029, 2023.

\bibitem{shin2024japanese_jsl1}
Jungpil Shin, Md~Al~Mehedi Hasan, Abu Saleh~Musa Miah, Kota Suzuki, and Koki Hirooka.
\newblock Japanese sign language recognition by combining joint skeleton-based handcrafted and pixel-based deep learning features with machine learning classification.
\newblock {\em CMES-COMPUTER MODELING IN ENGINEERING \& SCIENCES}, 2024.

\bibitem{10360810_miah_ksl2}
Jungpil Shin, Abu Saleh~Musa Miah, Kota Suzuki, Koki Hirooka, and Md. Al~Mehedi Hasan.
\newblock Dynamic korean sign language recognition using pose estimation based and attention-based neural network.
\newblock {\em IEEE Access}, 11:143501--143513, 2023.

\bibitem{miah2023dynamic_mcsoc}
Abu Saleh~Musa Miah, Jungpil Shin, Md~Al~Mehedi Hasan, Yuichi Okuyama, and Asai Nobuyoshi.
\newblock Dynamic hand gesture recognition using effective feature extraction and attention based deep neural network.
\newblock In {\em 2023 IEEE 16th International Symposium on Embedded Multicore/Many-core Systems-on-Chip (MCSoC)}, pages 241--247. IEEE, 2023.

\bibitem{miah2023skeleton_euvip}
Abu Saleh~Musa Miah, Jungpil Shin, Md~Al~Mehedi Hasan, Yusuke Fujimoto, and Asai Nobuyoshi.
\newblock Skeleton-based hand gesture recognition using geometric features and spatio-temporal deep learning approach.
\newblock In {\em 2023 11th European Workshop on Visual Information Processing (EUVIP)}, pages 1--6. IEEE, 2023.

\bibitem{wang2018temporal}
Limin Wang, Yuanjun Xiong, Zhe Wang, Yu~Qiao, Dahua Lin, Xiaoou Tang, and Luc Van~Gool.
\newblock Temporal segment networks for action recognition in videos.
\newblock {\em IEEE transactions on pattern analysis and machine intelligence}, 41(11):2740--2755, 2018.

\bibitem{simonyan2014very}
Karen Simonyan and Andrew Zisserman.
\newblock Very deep convolutional networks for large-scale image recognition.
\newblock {\em arXiv preprint arXiv:1409.1556}, 2014.

\bibitem{szegedy2015going}
Christian Szegedy, Wei Liu, Yangqing Jia, Pierre Sermanet, Scott Reed, Dragomir Anguelov, Dumitru Erhan, Vincent Vanhoucke, and Andrew Rabinovich.
\newblock Going deeper with convolutions.
\newblock In {\em Proceedings of the IEEE conference on computer vision and pattern recognition}, pages 1--9, Boston, MA, USA, 2015.

\bibitem{rahim2024advanced_miah}
Md~Abdur Rahim, Abu Saleh~Musa Miah, Hemel~Sharker Akash, Jungpil Shin, Md~Imran Hossain, and Md~Najmul Hossain.
\newblock An advanced deep learning based three-stream hybrid model for dynamic hand gesture recognition.
\newblock {\em arXiv preprint arXiv:2408.08035}, 2024.

\bibitem{radford2021learning}
Alec Radford, Jong~Wook Kim, Chris Hallacy, Aditya Ramesh, Gabriel Goh, Sandhini Agarwal, Girish Sastry, Amanda Askell, Pamela Mishkin, Jack Clark, et~al.
\newblock Learning transferable visual models from natural language supervision.
\newblock In {\em International conference on machine learning}, pages 8748--8763, Virtual Conference, 2021. PMLR.

\bibitem{lu2019vilbert}
Jiasen Lu, Dhruv Batra, Devi Parikh, and Stefan Lee.
\newblock Vilbert: Pretraining task-agnostic visiolinguistic representations for vision-and-language tasks.
\newblock {\em Advances in neural information processing systems}, 32, 2019.

\bibitem{li2019visualbert}
Liunian~Harold Li, Mark Yatskar, Da~Yin, Cho-Jui Hsieh, and Kai-Wei Chang.
\newblock Visualbert: A simple and performant baseline for vision and language.
\newblock {\em arXiv preprint arXiv:1908.03557}, 2019.

\bibitem{li2021supervision}
Yangguang Li, Feng Liang, Lichen Zhao, Yufeng Cui, Wanli Ouyang, Jing Shao, Fengwei Yu, and Junjie Yan.
\newblock Supervision exists everywhere: A data efficient contrastive language-image pre-training paradigm.
\newblock {\em arXiv preprint arXiv:2110.05208}, 2021.

\bibitem{li2022self}
Shuo Li, Fang Liu, and Licheng Jiao.
\newblock Self-training multi-sequence learning with transformer for weakly supervised video anomaly detection.
\newblock In {\em Proceedings of the AAAI Conference on Artificial Intelligence}, volume~36, pages 1395--1403, Vancouver, Canada, 2022.

\bibitem{lv2023unbiased}
Hui Lv, Zhongqi Yue, Qianru Sun, Bin Luo, Zhen Cui, and Hanwang Zhang.
\newblock Unbiased multiple instance learning for weakly supervised video anomaly detection.
\newblock In {\em Proceedings of the IEEE/CVF Conference on Computer Vision and Pattern Recognition}, pages 8022--8031, Vancouver, BC, Canada, 2023.

\bibitem{rahim2020hand}
Md~Abdur Rahim, Abu Saleh~Musa Miah, Abu Sayeed, and Jungpil Shin.
\newblock Hand gesture recognition based on optimal segmentation in human-computer interaction.
\newblock In {\em 2020 3rd IEEE International Conference on Knowledge Innovation and Invention (ICKII)}, pages 163--166. IEEE, 2020.

\bibitem{tiwari2023comprehensive_3DCNN}
Satyam Tiwari, Goutam Jain, Dasharathraj~K Shetty, Manu Sudhi, Jayaraj~Mymbilly Balakrishnan, and Shreepathy~Ranga Bhatta.
\newblock A comprehensive review on the application of 3d convolutional neural networks in medical imaging.
\newblock {\em Engineering Proceedings}, 59(1):3, 2023.

\bibitem{joo2023cliptsa}
Hyekang~Kevin Joo, Khoa Vo, Kashu Yamazaki, and Ngan Le.
\newblock Clip-tsa: Clip-assisted temporal self-attention for weakly-supervised video anomaly detection.
\newblock pages 3230--3234, Kuala Lumpur, Malaysia, 2023. IEEE, IEEE International Conference on Image Processing (ICIP).

\bibitem{zhu2023pdl}
Wenhui Zhu, Peijie Qiu, Oana~M. Dumitrascu, and Yalin Wang.
\newblock Pdl: Regularizing multiple instance learning with progressive dropout layers, 2023.

\bibitem{tvl1}
C.~Zach, T.~Pock, and H.~Bischof.
\newblock A duality based approach for realtime tv-l1 optical flow.
\newblock In {\em Proceedings of the 29th DAGM Conference on Pattern Recognition}, page 214–223, Berlin, Heidelberg, 2007. Springer-Verlag.

\bibitem{paszke2019pytorch}
Adam Paszke, Sam Gross, Francisco Massa, Adam Lerer, James Bradbury, Gregory Chanan, Trevor Killeen, Zeming Lin, Natalia Gimelshein, Luca Antiga, et~al.
\newblock Pytorch: An imperative style, high-performance deep learning library.
\newblock {\em Advances in neural information processing systems}, 32, 2019.

\bibitem{gollapudi2019learn}
Sunila Gollapudi.
\newblock {\em Learn computer vision using OpenCV}.
\newblock Springer, 2019.

\bibitem{dozat2016incorporating}
Timothy Dozat.
\newblock Incorporating nesterov momentum into adam.
\newblock In {\em Proceedings of the 4th International Conference on Learning Representations, Workshop Track}, pages 1--4, 2016.

\bibitem{zaheer2020self}
Muhammad~Zaigham Zaheer, Arif Mahmood, Hochul Shin, and Seung-Ik Lee.
\newblock A self-reasoning framework for anomaly detection using video-level labels.
\newblock {\em IEEE Signal Processing Letters}, 27:1705--1709, 2020.

\bibitem{wan2020weakly}
Boyang Wan, Yuming Fang, Xue Xia, and Jiajie Mei.
\newblock Weakly supervised video anomaly detection via center-guided discriminative learning.
\newblock In {\em 2020 IEEE international conference on multimedia and expo (ICME)}, pages 1--6, London, United Kingdom, 2020. IEEE.

\bibitem{majhi2021dam}
Snehashis Majhi, Srijan Das, and Fran{\c{c}}ois Br{\'e}mond.
\newblock Dam: dissimilarity attention module for weakly-supervised video anomaly detection.
\newblock In {\em 2021 17th IEEE International Conference on Advanced Video and Signal Based Surveillance (AVSS)}, pages 1--8, Virtual Conference, 2021. IEEE.

\bibitem{wu2021learning_mil}
Peng Wu and Jing Liu.
\newblock Learning causal temporal relation and feature discrimination for anomaly detection.
\newblock {\em IEEE Transactions on Image Processing}, 30:3513--3527, 2021.

\bibitem{yu2021cross}
Shenghao Yu, Chong Wang, Qiaomei Mao, Yuqi Li, and Jiafei Wu.
\newblock Cross-epoch learning for weakly supervised anomaly detection in surveillance videos.
\newblock {\em IEEE Signal Processing Letters}, 28:2137--2141, 2021.

\bibitem{feng2021mist}
Jia-Chang Feng, Fa-Ting Hong, and Wei-Shi Zheng.
\newblock Mist: Multiple instance self-training framework for video anomaly detection.
\newblock In {\em Proceedings of the IEEE/CVF conference on computer vision and pattern recognition}, pages 14009--14018, Nashville, TN, USA, 2021.

\bibitem{zaheer2022generative}
M~Zaigham Zaheer, Arif Mahmood, M~Haris Khan, Mattia Segu, Fisher Yu, and Seung-Ik Lee.
\newblock Generative cooperative learning for unsupervised video anomaly detection.
\newblock In {\em Proceedings of the IEEE/CVF conference on computer vision and pattern recognition}, pages 14744--14754, New Orleans, LA, USA, 2022.

\bibitem{zaheer2023clustering}
Muhammad~Zaigham Zaheer, Arif Mahmood, Marcella Astrid, and Seung-Ik Lee.
\newblock Clustering aided weakly supervised training to detect anomalous events in surveillance videos.
\newblock {\em IEEE Transactions on Neural Networks and Learning Systems}, 2023.

\bibitem{cao2023weakly}
Congqi Cao, Xin Zhang, Shizhou Zhang, Peng Wang, and Yanning Zhang.
\newblock Weakly supervised video anomaly detection based on cross-batch clustering guidance.
\newblock In {\em 2023 IEEE International Conference on Multimedia and Expo (ICME)}, pages 2723--2728, Brisbane, Australia, 2023. IEEE.

\bibitem{tan2024overlooked}
Weijun Tan, Qi~Yao, and Jingfeng Liu.
\newblock Overlooked video classification in weakly supervised video anomaly detection.
\newblock In {\em Proceedings of the IEEE/CVF Winter Conference on Applications of Computer Vision}, pages 202--210, Waikoloa, Hawaii, 2024.

\bibitem{yi2022batch}
Shuhan Yi, Zheyi Fan, and Di~Wu.
\newblock Batch feature standardization network with triplet loss for weakly-supervised video anomaly detection.
\newblock {\em Image and Vision Computing}, 120:104397, 2022.

\bibitem{gong2022multi}
Yiling Gong, Chong Wang, Xinmiao Dai, Shenghao Yu, Lehong Xiang, and Jiafei Wu.
\newblock Multi-scale continuity-aware refinement network for weakly supervised video anomaly detection.
\newblock In {\em 2022 IEEE International Conference on Multimedia and Expo (ICME)}, pages 1--6, Taipei, Taiwan, 2022. IEEE.

\bibitem{majhi2023human}
Snehashis Majhi, Rui Dai, Quan Kong, Lorenzo Garattoni, Gianpiero Francesca, and Francois Bremond.
\newblock Human-scene network: A novel baseline with self-rectifying loss for weakly supervised video anomaly detection.
\newblock {\em arXiv preprint arXiv:2301.07923}, 2023.

\bibitem{park2023normality}
Seongheon Park, Hanjae Kim, Minsu Kim, Dahye Kim, and Kwanghoon Sohn.
\newblock Normality guided multiple instance learning for weakly supervised video anomaly detection.
\newblock In {\em Proceedings of the IEEE/CVF Winter Conference on Applications of Computer Vision}, pages 2665--2674, Waikoloa, Hawaii, 2023.

\bibitem{sun2023long}
Shengyang Sun and Xiaojin Gong.
\newblock Long-short temporal co-teaching for weakly supervised video anomaly detection.
\newblock {\em arXiv preprint arXiv:2303.18044}, 2023.

\bibitem{wang2023attention}
Lin Wang, Xiangjun Wang, Feng Liu, Mingyang Li, Xin Hao, and Nianfu Zhao.
\newblock Attention-guided mil weakly supervised visual anomaly detection.
\newblock {\em Measurement}, 209:112500, 2023.

\end{thebibliography}

\begin{IEEEbiography}[{\includegraphics[width=1in,height=1.25in, clip,keepaspectratio]{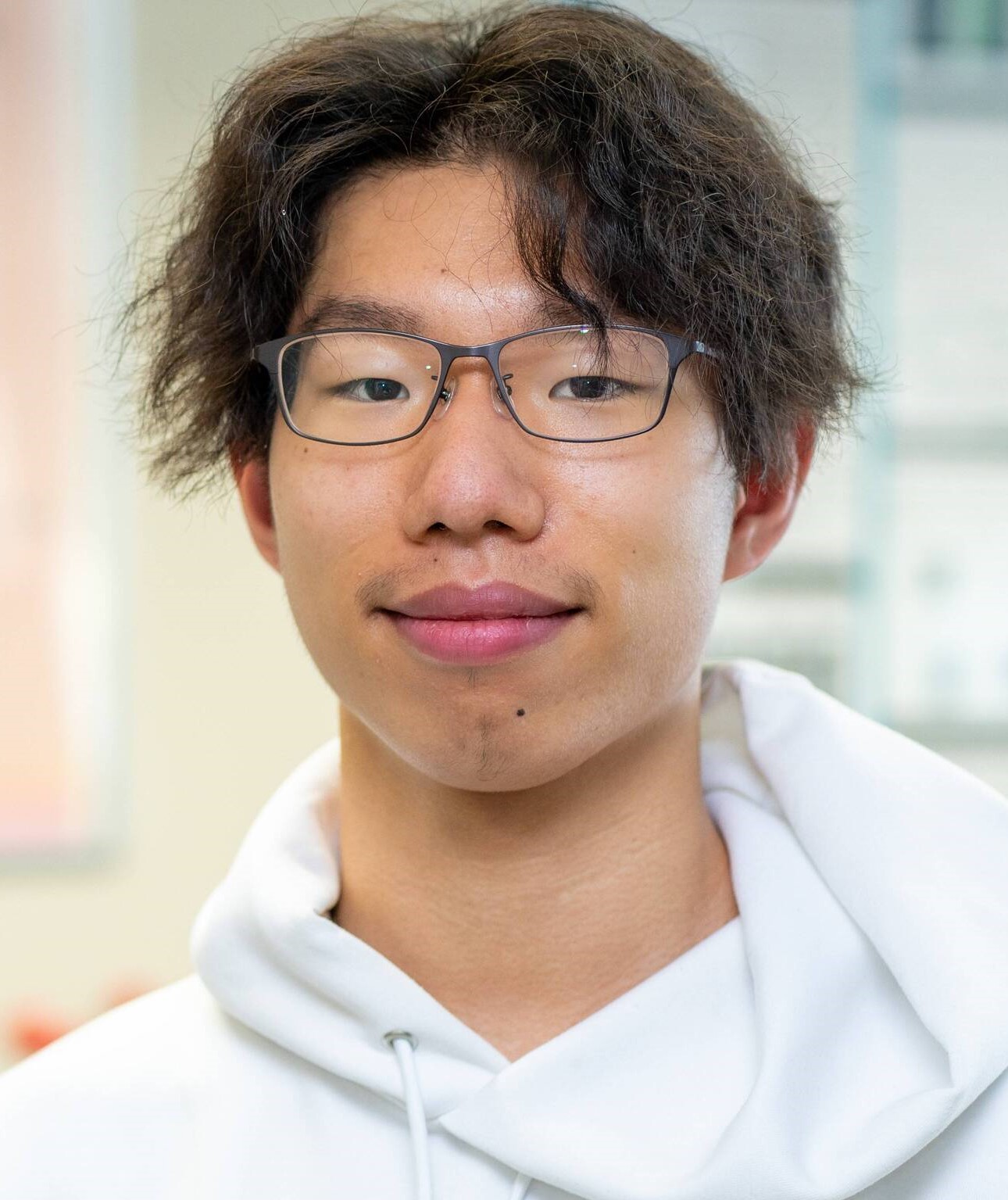}}]{Yuta Kaneko} is currently pursuing a bachelor’s degree in computer science and engineering with The University of Aizu (UoA), Japan. He joined the Pattern Processing Laboratory, UoA, in April 2023, under the direct supervision of Dr. Jungpil Shin. He is currently working on human activity recognition. His research interests include computer vision, pattern recognition, and deep learning.
\end{IEEEbiography}
\begin{IEEEbiography}[{\includegraphics[width=1in,height=1.25in,clip,keepaspectratio]{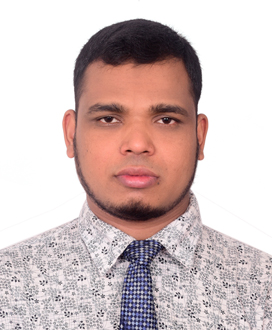}}]{Abu Saleh Musa Miah} (Member, IEEE) received the B.Sc.Engg. and M.Sc.Engg. degrees in computer science and engineering from the Department of Computer Science and Engineering, University of Rajshahi, Rajshahi-6205, Bangladesh, in 2014 and 2015, respectively, achieving the first merit position. He received his Ph.D. in computer science and engineering from the University of Aizu, Japan, in 2024, under a scholarship from the Japanese government (MEXT). He assumed the positions of Lecturer and Assistant Professor at the Department of Computer Science and Engineering, Bangladesh Army University of Science and Technology (BAUST), Saidpur, Bangladesh, in 2018 and 2021, respectively. Currently, he is working as a visiting researcher (postdoc) at the University of Aizu since April 1, 2024. His research interests include AI, ML, DL, Human Activity Recognition (HCR), Hand Gesture Recognition (HGR), Movement Disorder Detection, Parkinson's Disease (PD), HCI, BCI, and Neurological Disorder Detection. He has authored and co-authored more than 50 publications in widely cited journals and conferences.
\end{IEEEbiography}

\begin{IEEEbiography}[{\includegraphics[width=1in,height=1.25in,clip,keepaspectratio]{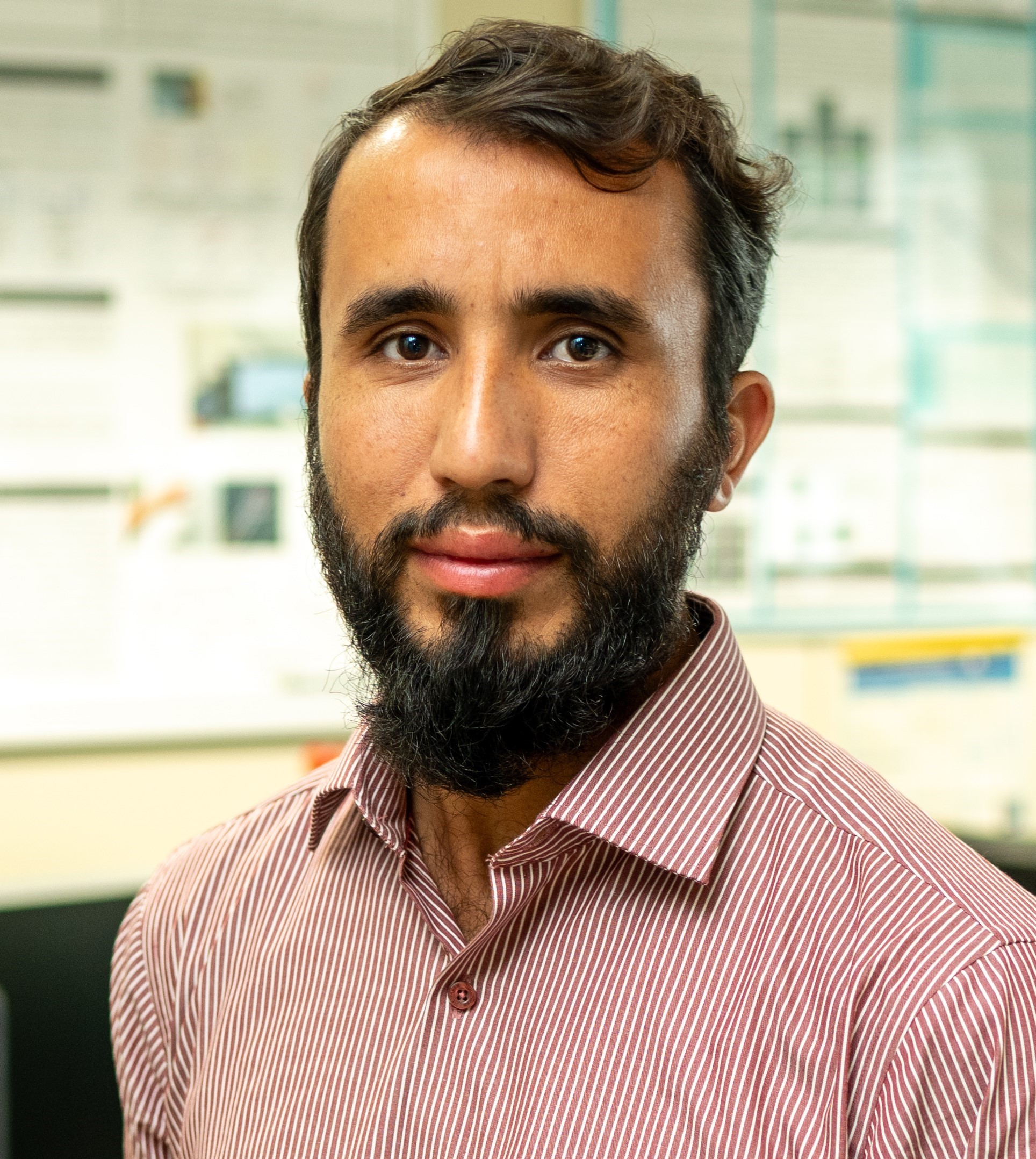}}]{Najmul Hassan} (Graduate Member, IEEE)  received the M.Sc. and MPhil degrees in Electronics from the University of Peshawar and Quaid e Azam University Islamabad Pakistan, in 2016 and 2018, respectively. He was a visiting researcher in the Department of Electronics, Quaid e Azam University Islamabad in 2022. He is currently pursuing a PhD degree with the School of Computer Science and Engineering, The University of Aizu, Japan, under a scholarship from the Japanese Government (MEXT) (since fall 2023). He has authored and co-authored more than seven publications published in widely cited journals and conferences. His main research interests are human action recognition, human gesture recognition, Alzheimer’s disease diagnosis, and image processing algorithms dealing with special images like underwater images, nighttime images, and foggy images. 
\end{IEEEbiography}


\begin{IEEEbiography}[{\includegraphics[width=1in,height=1.25in,clip,keepaspectratio]{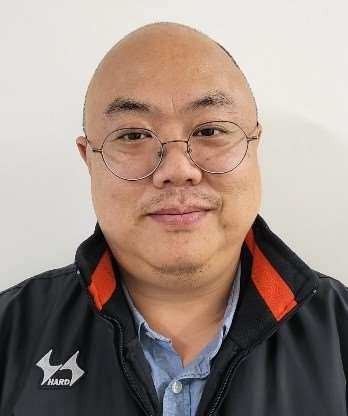}}] {HYUNG SUP LEE}  received the B.S., M.S., and Ph.D. degrees from Dong-Eui University, Pusan, South Korea, in 2004, 2006, and 2017, respectively. From 2012 to 2015, he served as CTO with Albam Company Ltd. Since 2014, he has been a Professor with the Department of Application Software Engineering at Dong-Eui University. His research interests include big data, data analysis, and artificial intelligence.
\end{IEEEbiography}

\begin{IEEEbiography}[{\includegraphics[width=1in,height=1.25in,clip,keepaspectratio]{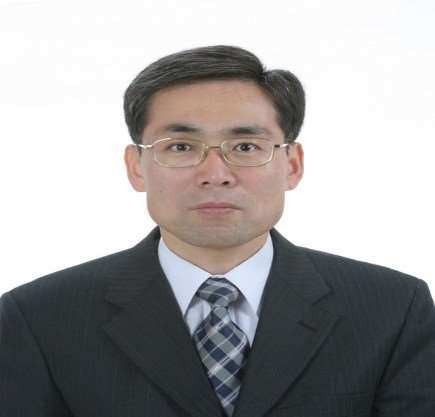}}] {SI-WOONG JANG} received the B.S., M.S., and Ph.D. degrees from Pusan National University, Pusan, South Korea, in 1984, 1993, and 1996, respectively. From 1986 to 1993, he was a Research Worker with Daewoo Telecom Company Ltd. Since 1996, he has been a Professor with the Department of Computer Engineering, Dong-Eui University, South Korea. His research interests include image processing, artificial intelligence, and the Internet of Things (IoT).
\end{IEEEbiography}

\begin{IEEEbiography}[{\includegraphics[width=1in,height=1.25in, clip,keepaspectratio]{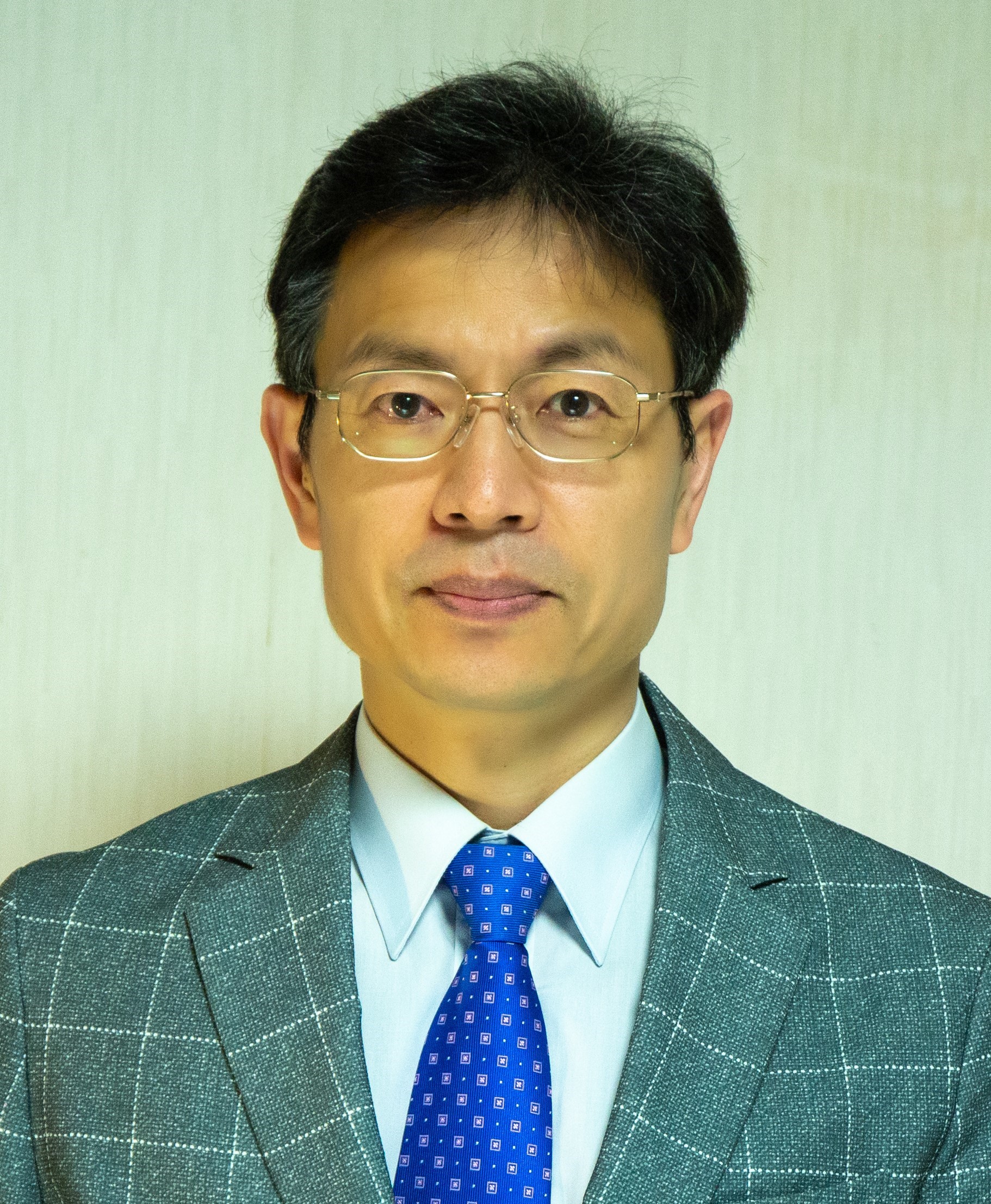}}]{Jungpil Shin} (Senior Member, IEEE) received a B.Sc. in Computer Science and Statistics and an M.Sc. in Computer Science from Pusan National University, Korea, in 1990 and 1994, respectively. He received his Ph.D. in computer science and communication engineering from Kyushu University, Japan, in 1999, under a scholarship from the Japanese government (MEXT). He was an Associate Professor, a Senior Associate Professor, and a Full Professor at the School of Computer Science and Engineering, The University of Aizu, Japan in 1999, 2004, and 2019, respectively. His research interests include pattern recognition, image processing, computer vision, machine learning, human-computer interaction, non-touch interfaces, human gesture recognition, automatic control, Parkinson’s disease diagnosis, ADHD diagnosis, user authentication, machine intelligence, bioinformatics, as well as handwriting analysis, recognition, and synthesis. He is an ACM, IEICE, IPSJ, KISS, and KIPS member. He served as program chair and as a program committee member for numerous international conferences. He serves as an Editor of IEEE Journals Springer, Sage, Taylor \& Francis, MDPI Sensors and Electronics, and Tech Science. He serves as an Editorial Board Member of Scientific Reports. He serves as a reviewer for several major IEEE and SCI journals. He has co-authored more than 400 published papers for widely cited journals and conferences.
\end{IEEEbiography}

\EOD
\end{document}